\documentclass[final,5p,times,twocolumn]{elsarticle}



\usepackage{amssymb}
\usepackage{tablefootnote}
\usepackage{bm}
\usepackage{booktabs}
\usepackage{amsmath}
\usepackage{multirow}
\usepackage{array}
\usepackage{tabularx}
\usepackage{setspace}
\usepackage{array}
\usepackage{makecell}
\newcolumntype{Y}{>{\RaggedRight\arraybackslash}X}

\usepackage{float}
\usepackage{hyperref,color}
\hypersetup{colorlinks=true,
linkcolor=black,
anchorcolor=red,
citecolor=cyan,
filecolor=red,
urlcolor=red,
pdfauthor=author}
\newcommand{\revc}[1]{\textcolor{black}{#1}}

\usepackage{lineno}

\makeatletter
\def\ps@pprintTitle{%
  \let\@oddhead\@empty
  \let\@evenhead\@empty
  \let\@oddfoot\@empty
  \let\@evenfoot\@empty
}
\makeatother

\begin{document}

\begin{frontmatter}



\title{Bridging Dual Knowledge Graphs for Multi-Hop Question Answering in Construction Safety}

\author[a]{Yuxin Zhang}\ead{yuxin0922@tamu.edu}
\author[a]{Xi Wang}\ead{xiwang@tamu.edu}
\author[a]{Mo Hu}\ead{mohu@tamu.edu}
\author[a]{Zhenyu Zhang\corref{cor1}}\ead{z.zhang@tamu.edu}
\cortext[cor1]{Corresponding author}
\affiliation[a]{organization={Department of Construction Science, College of Architecture, Texas A\&M University, College Station},
            country={USA}}

\begin{abstract}
\setlength{\parindent}{0pt}
Information retrieval and question answering from safety regulations are essential for automated construction compliance checking but are hindered by the linguistic and structural complexity of regulatory text. Many queries are multi-hop, requiring synthesis across interlinked clauses. To address the challenge, this paper introduces BifrostRAG, a dual-graph retrieval-augmented generation (RAG) system that models both linguistic relationships and document structure. The proposed architecture supports a hybrid retrieval mechanism that combines graph traversal with vector-based semantic search, enabling large language models to reason over both the content and the structure of the text. On a multi-hop question dataset, BifrostRAG achieves 92.8\% precision, 85.5\% recall, and an F1 score of 87.3\%. These results significantly outperform vector-only and graph-only RAG baselines, establishing BifrostRAG as a robust knowledge engine for LLM-driven compliance checking. The dual-graph, hybrid retrieval mechanism presented in this paper offers a transferable blueprint for navigating complex technical documents across knowledge-intensive engineering domains.

\vspace{1em}

\textbf{Note:} This is an accepted manuscript of an article published in \textit{Automation in Construction}. The final authenticated version is available online at \url{https://doi.org/10.1016/j.autcon.2026.106794}.

\end{abstract}

%


\end{frontmatter}



\section{Introduction}
Construction is one of the most hazardous industries in the United States, accounting for nearly 20\% of work-related fatalities across all industries \citep{chen2023knowledge}. While many incidents are preventable through proper safety enforcement, increasingly complex regulations present significant implementation challenges \citep{lu2015ontology,choe2016analysis}. Safety standards have grown more intricate due to \revc{an} expanding understanding of hazards and the continuous development of new safety equipment \citep{michaels2020occupational}. This regulatory complexity particularly affects workers with limited literacy, who struggle to comprehend dense safety documents \citep{schulte2003information}. To address this accessibility gap, knowledge retrieval and question-answering (QA) systems have emerged as promising solutions to simplify and clarify safety regulations for construction stakeholders \citep{solihin2016knowledge,zhou2023facilitating}. These QA systems also serve as critical components in automated field compliance checking systems \citep{wang2024few,wu2021combining,kulinan2024advancing}. Current compliance technologies integrate computer vision with natural language processing (NLP): first identifying objects and site conditions from field imagery, then mapping these observations to relevant regulatory requirements \citep{cui2025beyond,zhang2025dynamic}. With the recent advance of generative artificial intelligence (AI), QA systems are into incorporating retrieval-augmented generation (RAG) techniques, which enhance large language models (LLMs) by grounding their outputs in external regulatory knowledge \citep{lee2024performance}. The effectiveness of this compliance assessment pipeline fundamentally depends on the RAG integration that can accurately retrieve and interpret applicable safety rules.

A fundamental challenge in regulatory information retrieval stems from the linguistic and structural complexity of safety regulations. In addition to being rich in specialized domain knowledge, regulatory texts often feature multi-hop conditional logic and exceptions, further compounding their interpretive difficulty \citep{solihin2016knowledge,zhou2023facilitating}. Multi-hop questions are queries that require retrieving and synthesizing multiple interdependent pieces of information, often distributed across different sections or documents. For example, the Occupational Safety and Health Act for Construction (OSHA 1926) often states a general applicability in one provision while placing exceptions or complementary information in entirely different sections. The manifestation of such relationship can vary considerably: some are explicitly indicated through direct citations or hierarchical referencing, whereas others remain implicit, emerging from shared terminology or logical inference without being clearly stated in the text \citep{chen2024knowledge} (See examples from Table~\ref{t2} in Section \ref{s22}). This structure necessitates QA systems capable of handling multi-hop questions by effectively locating, retrieving, and integrating dispersed information \citep{cui2025beyond}. Identifying these latent connections is crucial for reliable regulatory interpretation and compliance guidance.

However, identifying and translating regulatory relationships into machine-readable formats for multi-hop QA remains limited. Only a few studies have approached this challenge: Lee et al. developed a keyword-based approach for Korean OSHA standards, linking sections through shared terminology to facilitate clustered retrieval \citep{lee2024performance}. Advancing this concept, Chen et al. created an algorithmic method for water conservancy construction safety regulations that can filter out weak connections between keywords, thereby forming robust information networks from previously scattered document content \citep{chen2024knowledge}. In addition, Wu et al. developed a RAG-LLM system that structures construction data into a hierarchical knowledge tree to capture parent-child relationships, contextualizing child sections within broader textual frameworks \citep{wu2025retrieval}. However, significant knowledge gaps persist. Existing methods inadequately address cross-referencing scenarios and implicit relationships. Furthermore, while researchers have examined some relationship types in isolation, the field lacks a comprehensive framework for addressing the various relationship patterns within safety regulations. Prior research only focuses on single-hop questions or lack clear evidence of multi-hop question assessment. This leaves unexplored the systems' ability to navigate scenarios requiring multi-hop information retrieval across provisions connected through cross-references, shared terminologies, or relevant scopes.

To address these limitations, this \revc{paper} developed BifrostRAG\footnotemark, \footnotetext{\revc{The framework is named} ``BifrostRAG'' after Bifrost, the rainbow bridge in Norse mythology that connects distinct realms. This nomenclature reflects \revc{the} framework's core function: bridging fragmented safety regulations through dual knowledge graphs and hybrid retrieval to synthesize diverse information. The name underscores the architectural duality and hybrid mechanism that distinguish \revc{this} approach from existing single-graph RAG systems.}a dual knowledge graph-enhanced RAG system that captures both explicit and implicit relationships between construction safety rules. BifrostRAG integrates two complementary knowledge graphs to map hierarchical structures and cross references while also connecting conceptually related provisions. Knowledge graphs provide an intuitive abstraction of regulatory entities and their relationships, enabling efficient traversal across interconnected information when queried \citep{guo2022automatic,hogan2021knowledge}. This capability makes knowledge graphs particularly promising for addressing knowledge-intensive tasks that require integrating multiple regulatory provisions \citep{dong2023hierarchy}. Notably, BifrostRAG integrates LLMs into the pipeline to automate critical tasks, including entity extraction, relation identification, and mapping of regulatory relationships. While researchers have begun exploring using deep learning models to construct knowledge graphs for construction compliance, they focus on linguistic complexities at the sentence level, such as resolving ambiguities and capturing non-verbal relations \citep{ren2021semantic,wang2023deep,wang2023bdeep,wang2024information}. BifrostRAG addresses not only these sentence-level challenges, but also the higher-level structural integration needed to navigate connections between regulatory provisions. BifrostRAG is specifically designed to accommodate the unique structural and linguistic complexities of regulatory documents. \revc{The proposed} zero-shot LLM approach eliminates the need for manually crafted ontologies and specialized training processes, making the system more adaptable and maintainable.

In \revc{the} evaluation using 93 multi-hop questions, BifrostRAG significantly outperformed two state-of-the-art alternatives: a general-purpose knowledge graph RAG system (Neo4j) and OpenAI's default vector-based RAG solution. These results confirm that handling regulatory relationships in safety standards remains challenging for even advanced natural language retrieval techniques. To address these critical challenges, this paper introduces {BifrostRAG}, a RAG system featuring a dual knowledge graph architecture designed to explicitly model both explicit and implicit relationships within construction safety regulations. Leveraging a hybrid retrieval mechanism that combines structured graph traversal with semantic vector-based search, BifrostRAG enables LLMs to effectively retrieve and synthesize fragmented regulatory information for multi-hop question answering. \revc{This paper} provides a scalable solution that enhances regulation accessibility and regulatory compliance assessments for construction safety professionals.

\section{Research Background and Challenges}
\revc{This section reviews the OSHA 1926 regulatory document and analyzes its complexity, which poses substantial challenges for multi-hop regulatory QA. The identified multi-hop patterns within the document serve as a conceptual foundation for the subsequent literature review, system design, and evaluation.}
\subsection{OSHA Standards Naming and Numbering System}
\begin{figure}[h]
\centering
\includegraphics[width=\columnwidth]{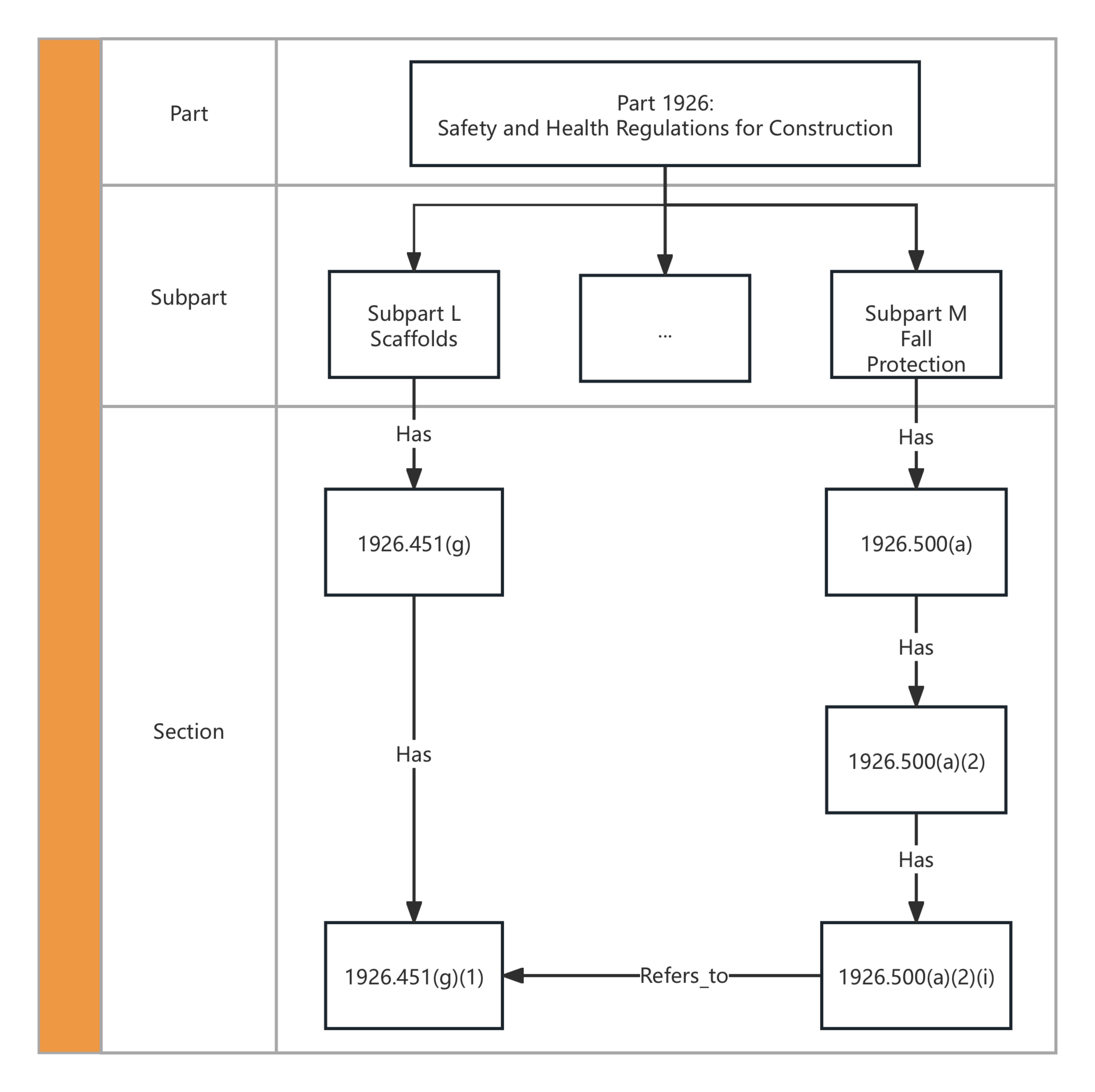}
\caption{Structure of OSHA 1926}
\label{f1}
\end{figure}

Occupational safety and health surveillance is recognized as a critical determinant influencing the successful execution of construction projects. In the United States, this regulatory commitment is exemplified through OSHA Part 1926--the Safety and Health Regulations for Construction Act. These regulations encapsulate a multifaceted approach to workplace safety, including hazard identification and prevention, site inspections, training, planning, accident investigations and reporting procedures, and safety performance. OSHA Part 1926 follows a systematic hierarchical structure with standardized naming and numbering conventions, as illustrated in Figure~\ref{f1}. The regulation organizes content into subparts, with each section using a parent-child framework that progresses from broad categories to specific requirements. For example, section 1926.500(a) branches into subsection 1926.500(a)(2), which further subdivides into 1926.500(a)(2)(i). This nested structure uses lowercase letters, Arabic numerals, and lowercase Roman numerals to maintain clear organizational flow. The regulation also incorporates extensive cross-references between f sections across different subparts, creating an interconnected network of regulatory requirements.

\subsection{Structure of Safety Regulations}\label{bg}
Safety regulatory frameworks, exemplified by OSHA 1926, constitute complex networks of interconnected provisions rather than collections of isolated rules. Compliance necessitates understanding both individual provisions and their relationships. Within OSHA 1926, four primary relationship patterns characterize provision interactions: exception, conditional, complementary, and cross-cutting relationships, as detailed in Table~\ref{t1}.

\begin{table*}[h]
\centering
\footnotesize
\linespread{0.9}\selectfont
\caption{Taxonomy of Relationship Types in OSHA 1926 Regulatory Framework}
\label{t1}
\begin{tabular}{>{\raggedright\arraybackslash}p{2cm}>{\raggedright\arraybackslash}p{5cm}>{\raggedright\arraybackslash}p{5cm}>{\raggedright\arraybackslash}p{2.7cm}}
\toprule
\textbf{Relationship} & \textbf{Description} & \textbf{Example} &\textbf{Sources}\\
\midrule
Exception & Provisions that establish specific exemptions or modifications to a general rule under defined circumstances & \S1926.451(b)(2) establishes that scaffold platforms must be at least 18 inches wide, except as provided in \S1926.451(b)(2)(i) and (ii), which permit narrower platforms in restricted areas. &OSHA, Letters of Interpretation (12/12/1996; 04/16/2002; 10/23/2002; 03/03/2010)\tablefootnote{ https://www.osha.gov/laws-regs/standardinterpretations/1996-12-12}\tablefootnote{
https://www.osha.gov/laws-regs/standardinterpretations/2002-04-16}\tablefootnote{ https://www.osha.gov/laws-regs/standardinterpretations/2002-10-23}\tablefootnote{
https://www.osha.gov/laws-regs/standardinterpretations/2010-03-03
}\\
\addlinespace
Conditional & Provisions whose requirements are activated or varied based on specific triggering conditions or scenarios & \S1926.451(g)(1)(i) through (vii) establish fall protection requirements based on different scaffold type&OSHA, Letters of Interpretation (09/14/2006
; 04/05/2005; 08/06/2002)\tablefootnote{https://www.osha.gov/laws-regs/standardinterpretations/2002-08-06-0}\tablefootnote{
https://www.osha.gov/laws-regs/standardinterpretations/2004-09-14}\tablefootnote{
https://www.osha.gov/laws-regs/standardinterpretations/2005-04-05
} \\
\addlinespace
Complementary & Multiple provisions addressing distinct aspects of the same safety objective, which must be applied concurrently for full compliance & \S1926.501(b)(4)(ii) requires holes to be protected by covers. \S1926.502(i) provides the necessary performance specifications. Both provisions are required for complete compliance. &OSHA, Letters of Interpretation (11/17/2004
; 12/22/2003)\tablefootnote{https://www.osha.gov/laws-regs/standardinterpretations/2003-12-22-1}\tablefootnote{
https://www.osha.gov/laws-regs/standardinterpretations/2004-11-17-0
}\\
\addlinespace
Cross-Cutting & Provisions related to different compliance objectives that must nevertheless be applied jointly in specific operational contexts due to overlapping scope & During heavy rain, a foreman must simultaneously apply \S1926.651(k)(1) for workplace inspections and \S1926.651(h) for water accumulation control, as both address different but concurrent safety concerns. &OSHA, Letters of Interpretation (09/30/1991; 03/23/1992; 08/10/2000)\tablefootnote{https://osha.oregon.gov/OSHARules/interps/im-94-24.pdf}\tablefootnote{
https://www.osha.gov/laws-regs/standardinterpretations/1992-03-23}\tablefootnote{
https://www.osha.gov/laws-regs/standardinterpretations/2000-08-10-0
}\\
\bottomrule
\end{tabular}
\end{table*}

These inter-provision relationships emerge through four underlying mechanisms: cross-reference, hierarchical structure, shared terminology, and logical necessity, with their description provided in Table~\ref{t2}. While cross-references and hierarchical structures provide explicit provision-to-provision mappings, many connections are implicit. Regulatory citations often create ambiguity when one provision refers broadly to an entire subpart without specifying a section. In these cases, provisions are linked conceptually through shared terminology or logical necessity rather than through structural elements or explicit references. The varied mechanisms expressing both explicit and implicit relationships present significant challenges for conventional NLP techniques. Further complicating identification efforts is the inconsistent writing style across different parts of the OSHA 1926 standard, which hinders pattern recognition and uniform interpretation. These variations in regulatory language create barriers to automated compliance assessment and highlight the need for more sophisticated analytical approaches.

\begin{table*}[h]\linespread{1.05}\selectfont
\centering
\footnotesize
\caption{Enabling Mechanisms for Regulatory Relationships in OSHA 1926}
\label{t2}
\begin{tabular}{>{\raggedright\arraybackslash}p{3cm}>{\arraybackslash}p{4cm}>{\arraybackslash}p{8cm}}
\toprule
\textbf{Mechanism} & \textbf{Description} & \textbf{Example} \\
\midrule
Cross-reference (explicit) & Explicit citations that create direct links between provisions, establishing unambiguous pathways for navigation and compliance. & \S1926.502(b)(15) on synthetic rope inspection directly references \S1926.502(b)(3) for strength criteria, creating a clear complementary relationship between the provisions. \\
\addlinespace
Hierarchical structure (explicit) & Formal organization where parent provisions establish general requirements or context, governing the application of more specific child provisions & \S1926.451(d)(10) serves as a parent provision establishing general inspection requirements for suspension scaffold ropes, while its child provisions (1926.451(d)(10)(i)-1926.451(d)(10)(vi)) specify when identified defects trigger replacement requirements. \\
\addlinespace
Shared terminology (implicit) & Provisions linked conceptually using common terms, even without direct citation & \S1926.651(e) mandates protect employees from falling materiel while \S1926.651(j)(2) details specifications for falling material prevention in excavation. Though not explicitly linked, these provisions are connected through shared terminology. \\
\addlinespace
Relevant scope (implicit) & Multiple regulatory provisions apply to the relevant set of activities, conditions, equipment, or workplaces. & \S1926.651(k)(1) requires inspections during and after rainstorms, while \S1926.651(h) outlines control measures for accumulated water. Although these provisions do not reference each other directly or share terminology, they share logically relevant condition: once a hazard is identified, appropriate safety controls are triggered \\
\bottomrule
\end{tabular}
\end{table*}

\subsection{Multi-hop Questions}
Due to the structural complexity of safety regulations, safety professionals frequently encounter questions that require integrating information from multiple, non-contiguous provisions—a process known as multi-hop QA \citep{mavi2024multi,cui2025beyond}. This process demands navigation through the relationship types and enabling mechanisms described in Table~\ref{t1} and \ref{t2}. To illustrate this multi-hop process, consider this safety question: ``Can a combination of a warning line system and personal fall arrest system be used for all types of leading-edge work?'' Answering this question requires traversing several provisions through interconnected steps:

\begin{itemize}
\item First hop: \S1926.501(b)(2)(i) establishes the fundamental requirements for leading edge work conducted 6 feet or more above lower levels, notably excluding warning line systems from permitted fall protection options.
\item Second hop: Through an explicit cross-reference, \S1926.501(b)(2)(i) directs to \S1926.502(k), which allows for alternative fall protection plans when standard systems prove infeasible, though still not permitting warning line systems.
\item Third hop: \S1926.501(b)(10), which shares terminology with \S1926.501(b)(2)(i), permits ``a combination of warning line system and personal fall arrest system,'' but applies conditionally only to ``roofing work on low-slope roofs.''
\item Fourth hop: \S1926.501(b)(12), which shares terminology with \S1926.501(b)(2)(i) and (b)(10), addresses precast concrete erection with requirements similar to leading edge work, yet maintains the prohibition on warning line systems.
\end{itemize}

This example demonstrates how answering a one-to-many-reference, multi-hop question requires navigating a complex web of conditional requirements and exceptions across different regulatory contexts. The process involves both explicit cross-references and implicit connections through shared terminology, illustrating the intricate interplay between various relationship types and enabling mechanisms in regulatory interpretation.

\section{Literature Review}
\revc{This section reviews recent advances in knowledge graphs, RAG, and LLM-enabled regulatory QA. It emphasizes the limitations of existing representative approaches in handling complex, multi-hop queries.}
\subsection{Knowledge Graph}
The complex, multi-hop nature of regulatory QA suggests the need for more structured representation of regulatory knowledge. Knowledge graphs offer a promising solution by explicitly modeling relationships between regulatory provisions as formalized, machine-readable structures. A knowledge graph represents interlinked entities--objects, concepts, and events--within a graph-structured data network where relationships are explicitly encoded \citep{ehrlinger2016towards,paulheim2017knowledge}. This structure provides an intuitive framework for efficient knowledge retrieval \citep{hogan2021knowledge}. Since the advent of Semantic Web technologies, the term \textit{knowledge graph} has gained significant traction in both academic research and industry \citep{malyshev2018getting,yahya2021semantic}. It is widely employed in applications such as search engines \citep{zou2020survey} and social media platforms \citep{qian2019social,wang2018deep}. They are particularly valuable for multi-hop QA systems due to their structured representation and traceable information pathways.

Knowledge graphs store information in specialized graph databases that optimize management of graph-structured data. These systems represent information as nodes (entities) connected by edges (relationships). Users retrieve information by formulating queries in languages such as SPARQL or Cypher that specify relationship patterns, typically following a triple format: (head entity, relation, tail entity). For example, Zhu et al. demonstrated this query-based approach by mapping a knowledge graph onto BIM data to identify quality compliance violations \citep{zhu2025research}.

Knowledge graph construction traditionally relies on extensive manual effort guided by ontologies--predefined rules specifying entity categories and relationships within hierarchical structures \citep{kommineni2024human}. However, knowledge graph construction remains a challenging process due to the integration of large volumes of heterogeneous metadata, ambiguity in text, complex nested entities, and the presence of noise such as homonyms. Traditional methods often resort to human experts to manually design ontologies, populate them with instances, and validate factual accuracy while ensuring traceability of data provenance \citep{kommineni2024human}. Apart from the substantial time and cost incurred, the traditional methods are more vulnerable to subjective interpretations. Moreover, such approaches can also significantly compromise completeness and consistency, leading to missing critical facts. To address these limitations, recent efforts have shifted toward machine learning-based techniques \citep{asprino2023knowledge,ji2022survey,zhong2023comprehensive}, with statistical models, convolutional neural networks, and transformer architectures demonstrating promising capabilities in resolving linguistic ambiguities, interpreting numeric or conditional thresholds, and balancing granularity in entity and relation extraction \citep{al2020named,wang2023deep}. One of the most promising tools for efficiently constructing knowledge graphs is LLMs.

\subsection{LLMs and Knowledge Graph}\label{s22}
LLMs are transformer-based neural networks pre-trained on vast corpora of text, enabling them to understand and generate natural language with fluency and contextual awareness \citep{nassiri2023transformer}. However, it tends to face problems such as information hallucinations and catastrophic forgetting when it encounters domain-specific questions, where the RAG has become a powerful solution for fetching relevant documents from external knowledge sources and makes them available in the LLMs' input prompt \citep{huang2025survey,korbak2022reinforcement,liu2025chatqa,mcintosh2024culturally}. Given their advanced linguistic and reasoning capabilities, researchers are increasingly exploring how LLMs can support knowledge graph applications with three integration strategies emerging in recent literature \citep{pan2024unifying}.

The first approach, LLM-augmented knowledge graphs, leverages LLMs as sophisticated text processors to support knowledge graph construction and querying. In this model, LLMs identify entities, relationships, and attributes across text corpora and map them into structured graph formats \citep{kommineni2024human}. For example, Zhu et al. found that GPT-4 outperforms fine-tuned models in knowledge graph construction while reducing manual annotation work \citep{zhu2024llms}. For query generation, Meyer et al. demonstrated LLMs can effectively generate syntactically correct SPARQL queries from natural language questions, making knowledge graphs more accessible to non-technical users \citep{meyer2024assessing}.

The second approach, knowledge graph-enhanced LLMs, addresses an inherent limitation of LLMs: hallucination, or the generation of plausible but factually incorrect information. RAG offers a solution by supplementing LLMs with externally retrieved information \citep{huang2025survey,korbak2022reinforcement,liu2025chatqa,mcintosh2024culturally}. While traditional RAG systems rely on vector embeddings, knowledge graphs provide additional advantages through their explicitly structured relationships \citep{taipalus2024vector}, which are particularly valuable for multi-hop QA tasks.

The third approach involves a fully synergistic integration where LLMs and knowledge graphs enhance each other \citep{pan2024unifying}. In this paradigm, knowledge graphs provide grounded factual knowledge to detect and correct potential hallucinations, while LLMs offer nuanced language understanding for autonomous knowledge graph construction and maintenance \citep{wan2025empowering,pan2025context,zhang2025knowledge}. This bidirectional enhancement creates systems capable of both factual accuracy and linguistic interpretation, which is an ideal combination for complex regulatory compliance applications.

\subsection{Research Gaps}
Current research leveraging knowledge graphs to navigate the intricate web of interconnections within regulatory documents for multi-hop QA remains limited. While numerous studies have applied knowledge graphs at the sentence or provision level, their primary aim has often been to resolve linguistic ambiguities or extract specific requirements within individual clauses \citep{ren2021semantic,wang2023deep,wang2023bdeep,wang2024information}, rather than mapping the critical structural relationships between provisions that dictate overall compliance logic.

Table~\ref{t3} summarizes what has been explored in information retrieval for construction documents and what remains unaddressed in the previous work. Although recent work leverages semantic embeddings to retrieve similar keywords or passages, few studies directly tackle the structural complexity of documents. Lee et al., working with Korean OSHA standards, proposed a method linking sections based on shared terminology and then ranking clusters based on interconnections \citep{lee2024performance}. Chen et al. tackled similar document fragmentation challenges but focused their efforts earlier in the process during entity relationship extraction \citep{chen2024knowledge}. They developed algorithmic filters that assess relationships based on frequency and predictive power, extracting only the most meaningful connections to reduce the noise problem typically found in knowledge graph applications.  Wu et al. addressed a different aspect of regulatory complexity by developing a hierarchical knowledge tree for construction management documents \citep{wu2025retrieval}. It effectively addressed limitations of vector-based RAG systems that segment documents based on fixed token limits--a practice that often separates parent-child provisions into different chunks, compromising contextual understanding and retrieval accuracy. Their hierarchical approach successfully preserved parent-child relationships, ensuring topical coherence. Despite these advances, existing studies address only two of four critical structural relationship types, failing to capture the complete taxonomy of relationships within safety regulations. Furthermore, their evaluations focus exclusively on single-hop questions or lack clear evidence of multi-hop question assessment. This leaves unexplored the systems' ability to navigate scenarios requiring multi-hop information retrieval across provisions connected through cross-references, shared terminologies, or relevant scopes.

The potential for synergistic integration of LLMs with knowledge graphs remains largely conceptual in the context of safety regulations, especially concerning multi-hop QA. Critical gaps persist, including a lack of thorough evaluation of LLM capabilities for accurately extracting entities and the diverse types of relationships (explicit and implicit) from nuanced regulatory text. A particularly significant barrier lies in interfacing with knowledge graphs via traditional query languages like SPARQL and Cypher. These demand exact syntactic and semantic matches between the query and the KG schema \citep{francis2018cypher}. Even minor discrepancies in entity or relationship naming can lead to query failure. This brittleness is especially problematic in domains like construction safety regulations, characterized by referential ambiguity and inconsistent terminology, making it difficult for non-expert users and LLMs to formulate correct queries. For instance, Meyer et al. demonstrated that while LLMs can generate syntactically valid SPARQL, they often fail to correctly utilize the predefined classes, properties, and prefixes specific to a given knowledge graph schema \citep{meyer2024assessing}. Despite recognition of this query formulation challenge, the literature offers few robust solutions.

In summary, existing research lacks (1) comprehensive frameworks capable of identifying, representing, and reasoning over the full spectrum of relationship types within safety regulations; (2) rigorous validation on the capacity of knowledge graph-based systems to handle multi-hop QA; and (3) effective strategies for integrating LLMs with knowledge graphs to overcome knowledge extraction limitations and the query precision bottleneck with empirical evidence. To address these limitations, this paper introduces an end-to-end methodology that leverages LLMs to address the challenges of navigating complex safety regulations. Moving beyond previous approaches in literature that rely solely on keyword matching or hierarchical structures, \revc{this} system integrates both explicit and implicit relationships into a comprehensive knowledge graph architecture, enabling sophisticated multi-hop question answering.

\begin{table*}[h]\linespread{0.9}\selectfont
\centering
\setlength{\tabcolsep}{4pt}
\footnotesize
\caption{Research Status of Information Retrieval in Construction Documents}\label{t3}
\begin{tabular}{>{\raggedright\arraybackslash}p{2cm} >{\raggedright\arraybackslash}p{3.3cm} >{\raggedright\arraybackslash}p{1.8cm} >{\raggedright\arraybackslash}p{1.8cm} >{\raggedright\arraybackslash}p{3cm} >{\raggedright\arraybackslash}p{2.2cm}}
\toprule
\multirow{3}{2.5cm}{\textbf{Work}} & \multirow{3}{3cm}{\textbf{Retrieval mechanism}} & \multicolumn{4}{c}{\textbf{Retrievable information types}} \\\addlinespace
\cmidrule(lr){3-6}
& & \textbf{Cross-reference} & \textbf{Hierarchical structure} & \textbf{Shared terminology} & \textbf{Relevant scope} \\\addlinespace
\midrule
Ren \& Zhang (2021)\,\cite{ren2021semantic} & Rule-based information extraction
& Not supported & Not supported & Partially supported by gazetteers and the CPDC ontology & Not supported \\\addlinespace
Wang \& El-Gohary (2023)\,\cite{wang2023bdeep} & Limited to KG construction; retrieval not addressed.
& Not applicable & Not applicable & Not applicable & Not applicable \\\addlinespace
Wang et al. (2024)\,\cite{wang2024few} & Limited to KG construction; retrieval not addressed.
& Not applicable & Not applicable & Not applicable & Not applicable \\\addlinespace
Lee et al. (2024)\,\cite{lee2024performance} & KG-based Cypher query match
& Not supported & Not supported & Exact matching on surface forms; no synonyms or morphological changes & Partially supported by title and keyword filtering \\\addlinespace
Chen et al. (2024)\,\cite{chen2024knowledge} & Limited to KG construction; retrieval not addressed.
& Not applicable & Not applicable & Not applicable & Not applicable \\\addlinespace
Wu et al. (2025)\,\cite{wu2025retrieval} & Vector-based semantic similarity
& Not supported & Partially supported by grouping bottom-level nodes & Supported by semantic embedding & Not supported \\\addlinespace
Meyer et al. (2024)\,\cite{meyer2024assessing} & KG-based SPARQL pattern match
& Not supported & Not supported & Exact matching on surface forms; no synonyms or morphological changes & Not supported \\\addlinespace
Francis et al. (2018)\,\cite{francis2018cypher} & KG-based Cypher query match
& Not supported & Not supported & Exact matching on surface forms; no synonyms or morphological changes & Not supported \\\addlinespace
Wan et al. (2025)\,\cite{wan2025empowering} & Vector-based semantic similarity
& Not supported & Not supported & Supported by semantic embedding & Partially supported by schema constraints \\\addlinespace
\bottomrule
\end{tabular}
\end{table*}

\section{Technical Approach}
\revc{This section presents the proposed system architecture and technical approach for multi-hop QA, including the construction of the dual graphs and the design of the hybrid retrieval mechanism.}
\subsection{System Overview}

\begin{figure}[H]
\centering
\includegraphics[width=\columnwidth]{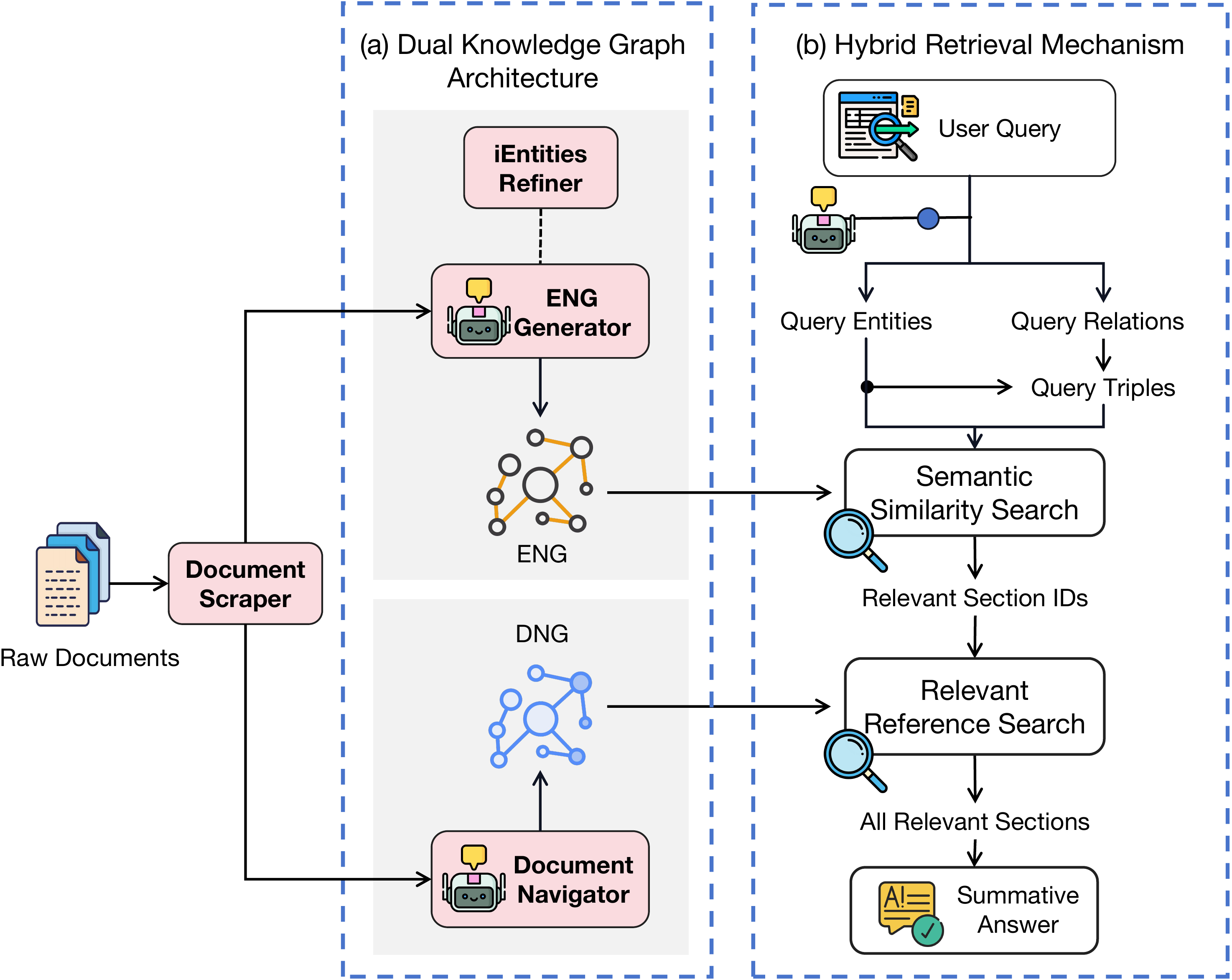}
\caption{ BifrostRAG Framework}
\label{f2}
\end{figure}

\revc{This paper} introduces BifrostRAG, a RAG-integrated system that enables LLMs to effectively reason over safety regulation documents characterized by semantic and structural complexity. BifrostRAG is built on a \textbf{dual knowledge graph architecture} (Figure~\ref{f2}a), which integrates two complementary components: the Entity Network Graph (ENG) and the Document Navigator Graph (DNG). The ENG captures semantic relationships between regulatory entities at the word and phrase level. Following preprocessing by the Document Scraper (Section \ref{s421}), an ENG Generator (Section \ref{s422}) constructs this graph by first extracting entities from regulatory text. An Incremental Entities Refiner (Section \ref{s423}) then consolidates semantically similar entities to reduce redundancy before the ENG Generator proceeds to extract relational triples (head entity–relation–tail entity). The resulting linguistic elements are encoded using an embedding model and stored in a vector database for semantic retrieval. The second graph DNG models structural relationships at the document level. A Document Navigator (Section \ref{s424}) builds this graph using section identifiers as nodes and establishing edges based on hierarchical relationships and cross-references within the regulations. The resulting graph structure is stored in a graph database for structural traversal across document sections. In this first phase, \revc{LLM agents are employed} to extract and validate the structural and semantic relationships from raw regulatory text and organize them into the DNG and ENG. This reflects LLM-enhanced knowledge graph construction.

These complementary graphs power BifrostRAG's \textbf{hybrid retrieval mechanism} (Figure~\ref{f2}b), which employs a two-stage retrieval process for QA. First, user questions are decomposed into entities and triples, then semantically matched against the ENG. This process identifies semantically relevant sections that share terminology and scope of application through entity and triple matching. Second, section IDs identified from the ENG become entry points for traversing the DNG, exploring parent sections for context, child sections for detail, and sibling sections for lateral relevance. The system ranks navigation paths and selects the most pertinent sections to generate the final grounded response. In this second phase, the constructed graphs serve as a structured external knowledge base to ground the LLM's reasoning, enabling it to generate precise, factual answers while mitigating hallucination. This reflects knowledge graph-enhanced LLM reasoning.

BifrostRAG demonstrates bidirectional integration between LLMs and knowledge graphs: LLMs facilitate entity extraction, graph construction, and question decomposition, while the knowledge graphs ground and contextualize LLM responses. In the remainder of this section, the dual knowledge graph generation process and hybrid retrieval mechanism are introduced in detail.

\subsection{Dual Knowledge Graph Generation}
The dual knowledge graph generation process is enabled by four modules: (1) Document Scraper for scraping and preprocessing raw regulatory text into a formatted dataset; (2) ENG Generator for ENG construction through entity and relationship extraction, including the identification of implicit links formed through shared terminology and logical necessity; (3) iEntities Refiner for similar entity consolidation; and (4) Document Navigator for DNG construction through hierarchical relationships and cross-references capture.

\subsubsection{Documents Scraper}\label{s421}
Authoritative documents such as OSHA standards are typically published in two formats: web-based and PDF. To convert these into an unencrypted and more processable format, the Document Scraper module employs two complementary methods, web scraping and PDF extraction. These methods can be used either independently or in combination to accurately compile the content of such documents.

Web scraping is the primary method for extracting content from web-based documents \citep{lee2024performance}. This involves sending an HTTP request with predefined headers to a specified list of URLs \citep{wu2025retrieval}. These requests retrieve the HTML source code of each webpage. The retrieved content is parsed using the Beautiful Soup library \citep{zhang2025dynamic}. This allows for the extraction of content bodies and their section numbers, which are then stored in a structured JSON format. To support multimodal learning in the future, the JSON structure also retains additional metadata, such as URLs and headings as references to images and tables.

PDF extraction distills plaintext layer from the PDF. The PDF files are first reframed by trimming the vertical and horizontal margins to remove headers, footers, and peripheral decorations such as logos, watermarks, and repetitive headlines. Some non-plaintext noises remained within the new page area especially images and tables are further removed using the PyMuPDF library. The remaining text fragments are concatenated into successive sentences.

When both web-based and PDF versions are available (e.g., OSHA standards), the two approaches can be used in combination to enhance the accuracy of the scraped content. Using Python built-in module difflib, a literal comparison is performed between the outputs of the web-scraped and PDF-extracted contents to ensure it is free from typographical errors or omissions. If the content from both sources aligns well, the verified data is stored in a structured SQL database for subsequent use. Otherwise, the discrepancies are manually reviewed, and only the correct version is retained.

\subsubsection{ENG Generator}\label{s422}
The ENG Generator serves as the core component of the proposed pipeline. An agentic workflow with multiple LLM agents collaborating to construct a knowledge graph is designed, including content pruning agent, entity extraction agent, relationship extraction, and ENG creation. Each agent is guided by a predefined system prompt that defines its role and general behavior, along with a flexible user prompt that adapts to the content of various text chunks (see the following details). This setup enables the agents to interact autonomously with data sources and perform specialized tasks independently within the system.

\begin{figure}
\centering
\includegraphics[width=\columnwidth]{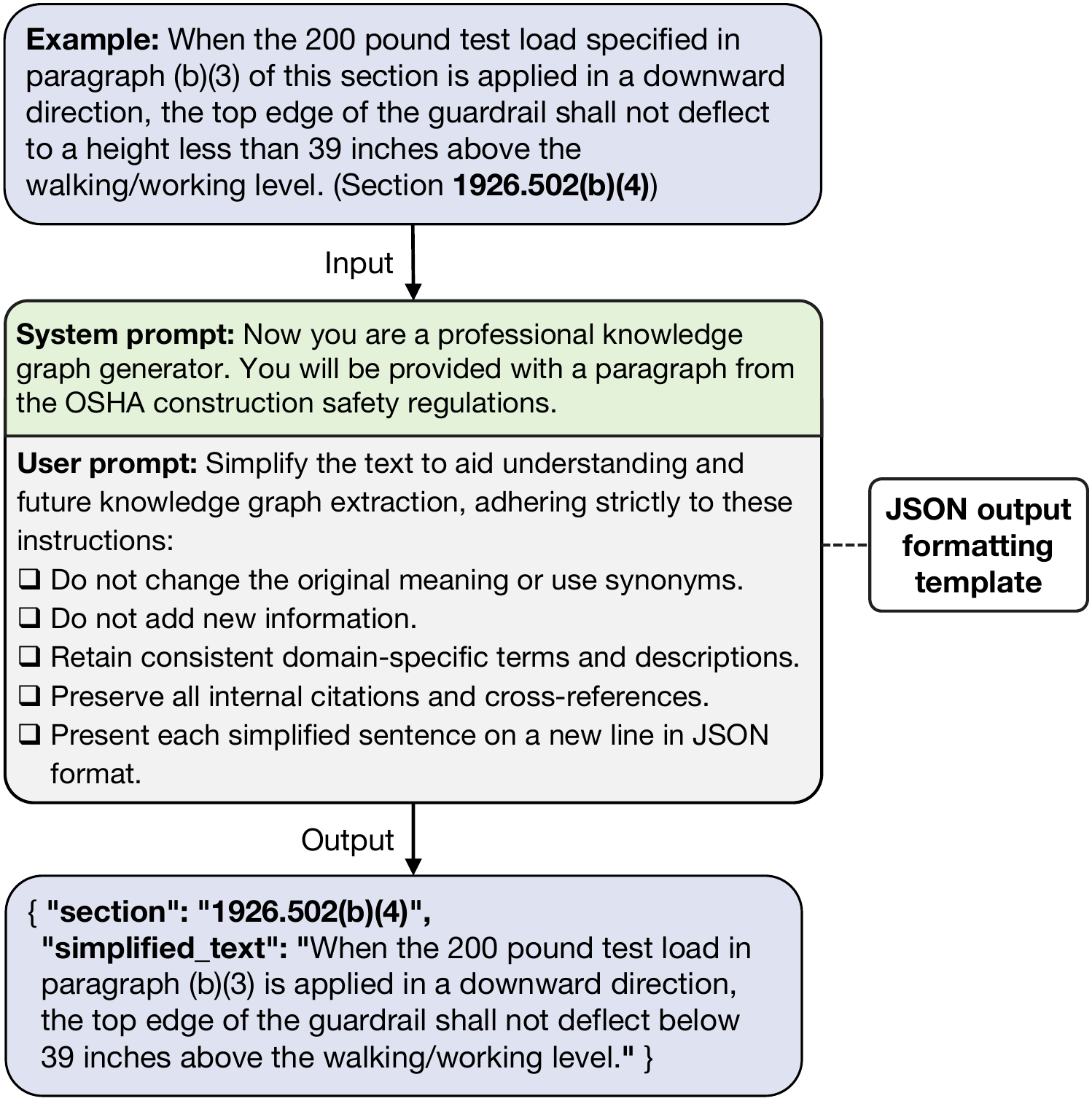}
\caption{Prompt Outline of Content Pruning}
\label{f3}
\end{figure}

\textbf{Content pruning:} The content pruning agent is tasked with simplifying textual content while carefully maintaining the sentence grammatical structures. The user prompt specifies comprehensive guidelines that the LLM agent must strictly follow during text processing. The agent is instructed to begin with a thorough review of the entire article before segmenting it into discrete paragraphs. Within each paragraph, the LLM simplifies sentences by preserving only their core structures and removing redundant or non-essential information, all while retaining the original meaning. Additionally, the agent is explicitly prohibited from substituting synonyms, thereby maintaining consistent terminology across the entire document. It must also retain all section IDs mentioned within the text to ensure accurate linkage with the Document Navigator Graph described in Section \ref{s424} For text fragments that are not complete sentences--typically section headings--the model is instructed to output the original text without modification. The final output must conform precisely to a predefined JSON format. An outline of the prompts is illustrated in Figure~\ref{f3}.

\begin{figure*}
\centering
\includegraphics[width=1\textwidth]{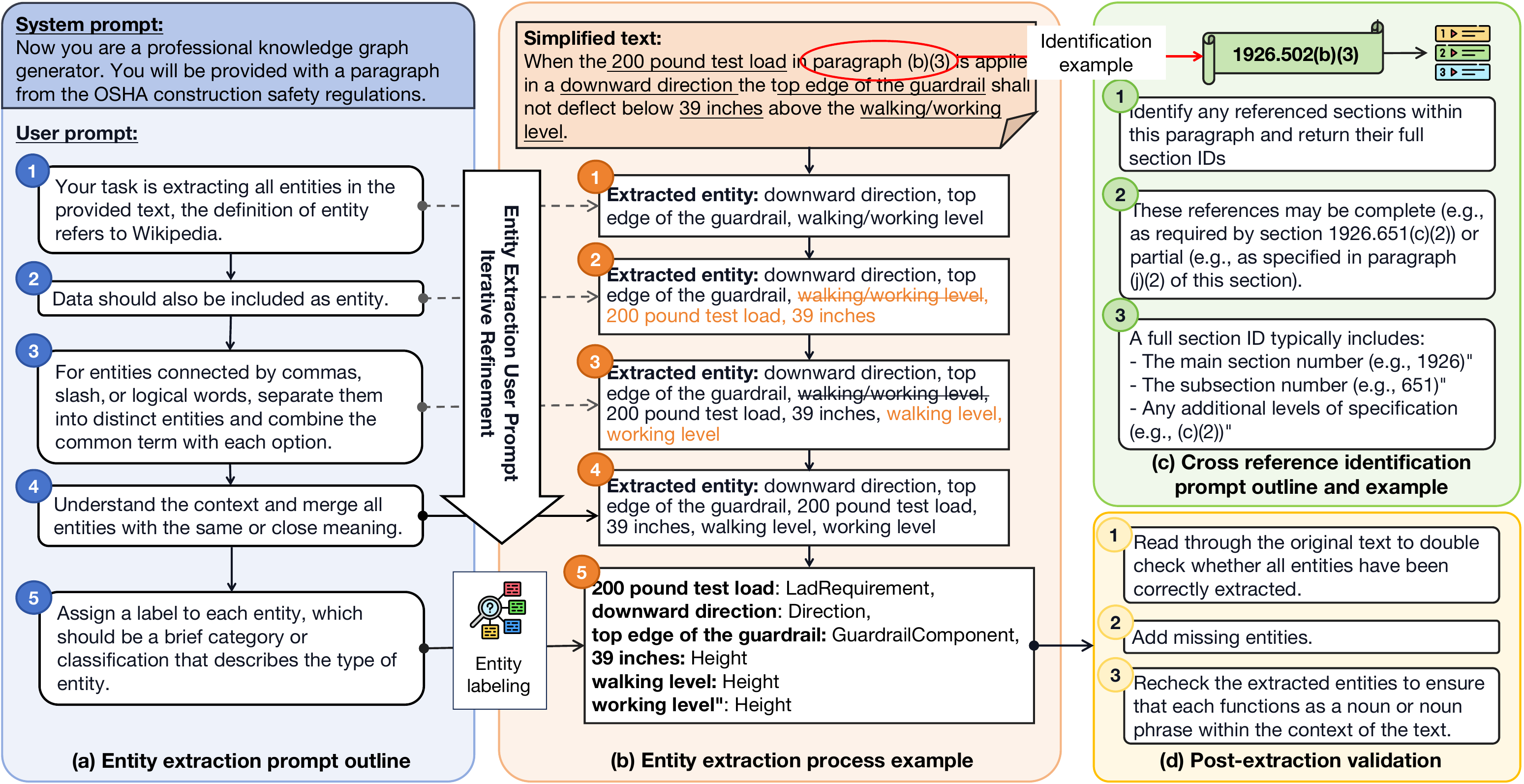}
\caption{Brief Prompt of Entity Extraction}
\label{f4}
\end{figure*}

\textbf{Entity Extraction:} The entity extraction agent integrates the simplified text produced by the content pruning agent into the user prompt as input. Then it requests LLM to extract entities in accordance with Wikipedia definitions, assigning representative labels based on prior knowledge. In addition to entity identification and labeling, the agent also extracts all referenced section IDs present in the input test, which is mandated through the prompt outlined in Figure~\ref{f4}c. This supports the construction of the knowledge graph in the Document Navigator Graph.

In order to achieve comprehensive yet precise entity extraction (i.e., captures all relevant entities while avoiding inclusion of irrelevant terms or phrases), the prompt design underwent iterative refinement during preliminary experiments. Adjustments were made to better align the agent's outputs with the manually annotated reference results. Through multiple rounds of testing and evaluation, several critical prompts were identified (Figure~\ref{f4}a), with an example of their corresponding impact on model performance shown in Figure~\ref{f4}b.

Following the initial entity extraction, a post-extraction validation process (Figure~\ref{f4}d) is introduced to scrutinize the quality of the extracted entities. Both the full original text and the previously extracted entities are integrated into the user prompt as the input. The prompts for this validation process are designed to detect and correct extraction errors, eliminating entities that deviate from the defined criteria and identifying any missing entities or labels. The objective of this step is to improve the overall accuracy and completeness of the resulting knowledge graph. An additional consideration is that although some words appear in nominal form, they do not function as nouns in the sentence. Hence, the agent must carefully interpret context and output only those entities that act as nouns within their specific linguistic context.

\begin{figure*}[h]
\centering
\includegraphics[width=0.9\textwidth]{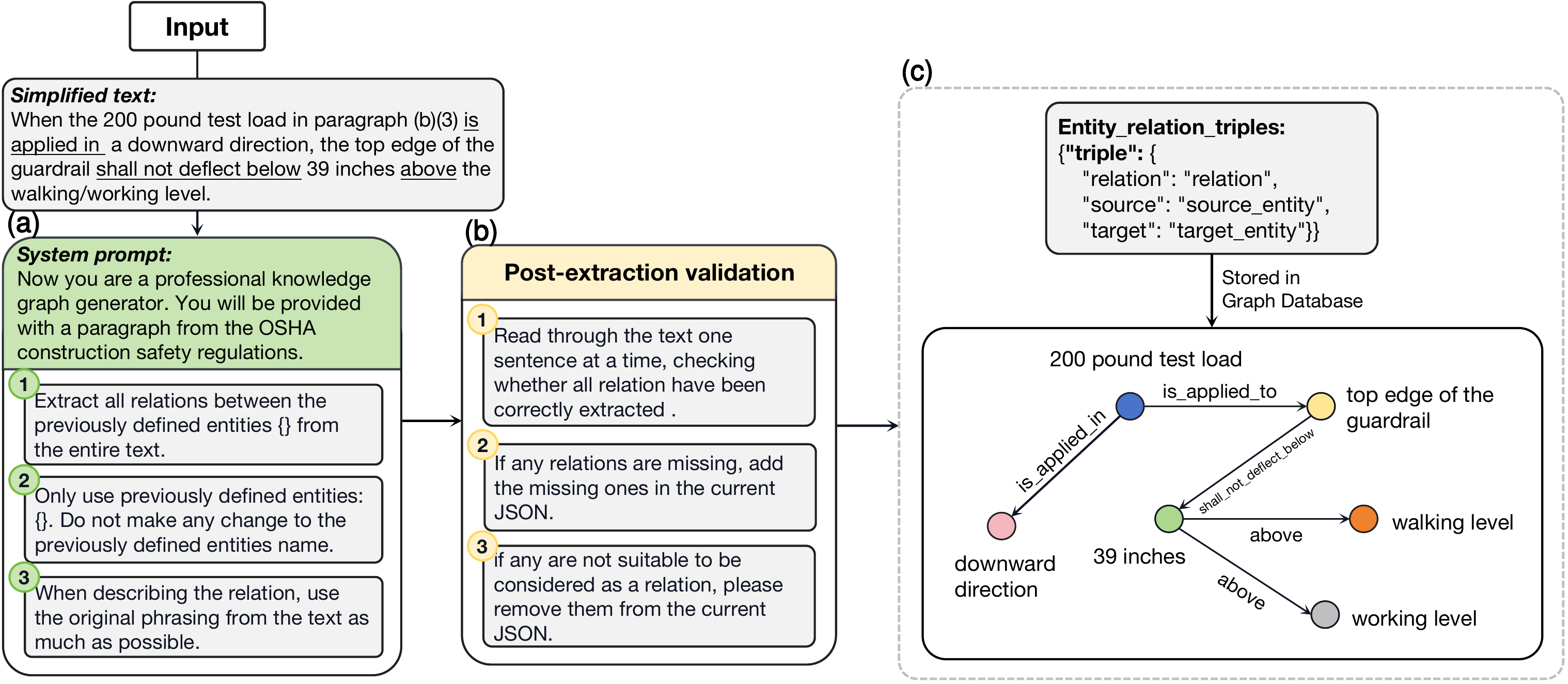}
\caption{ Brief Prompt of Relationship Extraction}
\label{f5}
\end{figure*}

\textbf{Relationship Extraction:} Following entity extraction, the semantic relationships among the identified entities are established, using another LLM agent (Figure~\ref{f5}a). A critical requirement in this process is that both the head and tail nodes must correspond strictly to previously identified entities. Furthermore, the relationship must be explicitly present in the text between those two entities. This constraint is essential to avoid the generation of fabricated relationships and to ensure all relationships are authentic (i.e., connect actual, verified entities). To maintain consistency and minimize discrepancies due to spelling or formatting differences, all extracted relationship phrases are standardized using snake\_case formatting.

Similar to entity extraction, a series of post-extraction validation is designed to rigorously assess the quality of the extracted relationships (Figure~\ref{f5}b). These inspections involve meticulous verification steps to confirm that each entity in the extracted triples matches the validated entities from the earlier step and that the relationship accurately reflects the grammatical and semantic structure of the original text, thereby ensuring the knowledge graph's integrity. Once both entities and relationships are thoroughly validated, complete triples (head entity, relationship, tail entity) are formulated and structured in a specified JSON format for input into the Neo4j Graph DBMS (Figure~\ref{f5}c).

The relationship extracts agent interfaces directly with the Neo4j Graph DBMS using a driver object authenticated via secure tokens to create the ENG. Utilizing Cypher query language, the system first creates nodes for each entity instance, assigning attributes including entity names, labels, global identifiers, and associated source section identifiers. Subsequently, additional Cypher commands are executed to establish relationships between these nodes, thereby embedding both the entities and their interconnections into Neo4j database structure.

\subsubsection{Incremental Entities Refiner (iEntities Refiner)}\label{s423}
To enhance the refinement process of the extracted entities, an iEntities Refiner is introduced, which systematically lemmatizes and consolidates entities exhibiting high semantic similarity. The process begins with lemmatizing nouns within each entity, converting them into more generalized forms, known as lemmas. Specifically, this step involves: 1) transforming plural nouns into their singular forms; 2) removing naming conventions, such as CamelCase (e.g., convert ``ExcavationWork'' to ``excavation work''); and 3) standardizing the spelling to lowercase (except for proper nouns). Each lemmatized entity is then matched against existing entries in the local schema, which maintains a structured and up-to-date record of entities and relationships currently existing within the graph database. If a match is found, the refiner updates the local schema by appending the corresponding section identifier to the matched entity property. Otherwise, the process advances to the next step.

\begin{figure*}[h]
\centering
\includegraphics[width=0.8\textwidth]{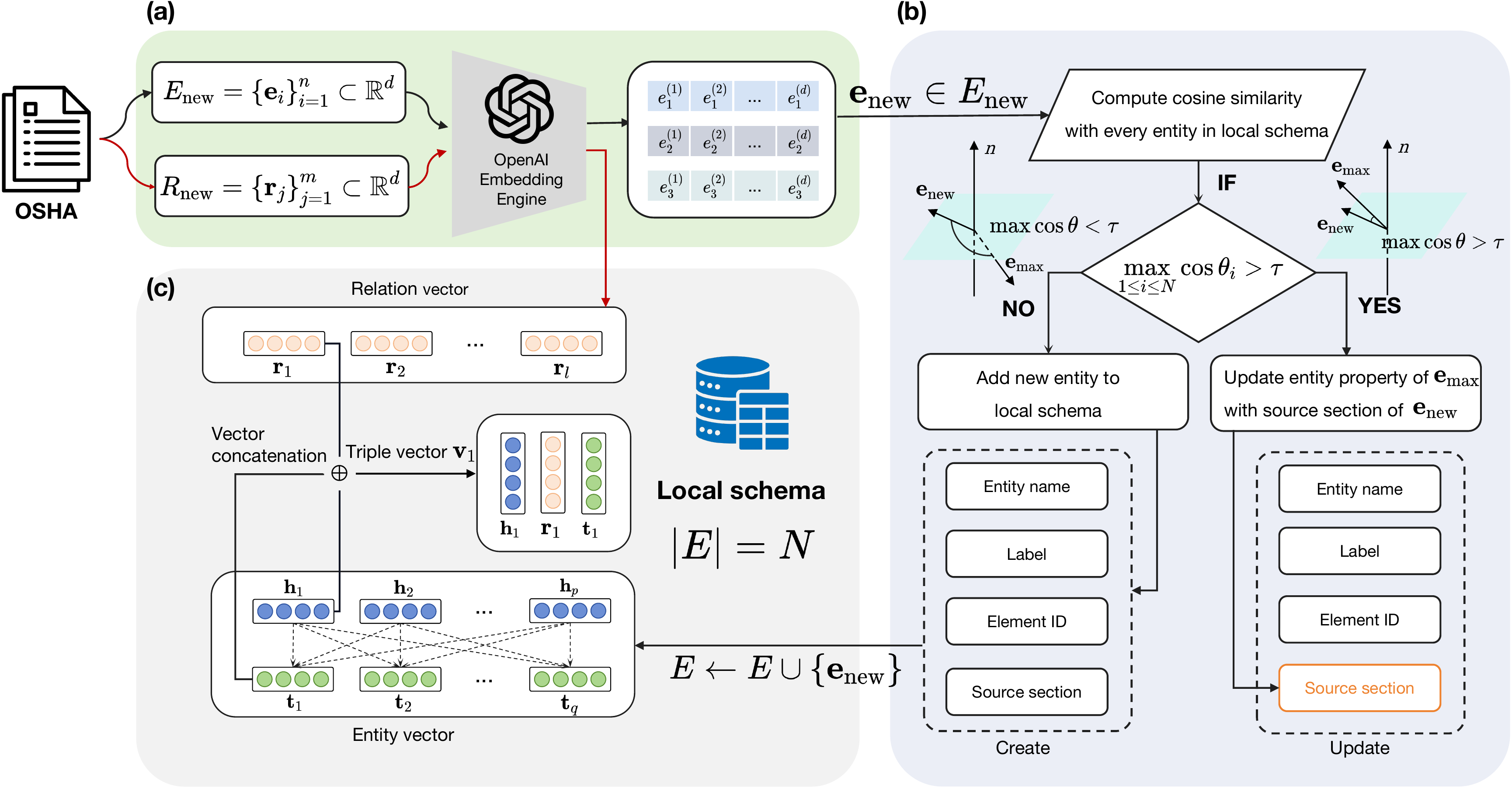}
\caption{Embedding-Based Incremental Entity Refinement}
\label{f6}
\end{figure*}

To further reduce duplication and improve entity matching accuracy, vector embeddings are incorporated. As shown in Figure~\ref{f6}a, both the newly extracted entity strings $E_{\text{new}}$ and the relation strings $R_{\text{new}}$ are transformed into vector representations via the embedding engine. The relation vectors are directly collected into the local schema, whereas for each new entity $\mathbf{e}_{\text{new}}
$, the iEntities Refiner performs an iterative cosine similarity computation between its embedding and the set of all existing entity embeddings $E$ maintained in the local schema, let $\mathbf{e}_{k}$ denote an arbitrary entity belongs to $E$. Assume that there are $N$ entities in local schema,
\begin{equation}
\forall\ \mathbf{e}_{k}\in E,\quad\cos\theta_{k}=\frac{\mathbf{e}_{\text{new}}\cdot\mathbf{e}_{k}}{\|\mathbf{e}_{\text{new}}\|\,\|\mathbf{e}_{k}\|},\quad k=1,\ldots,N.
\label{e1}
\end{equation}
As presented in Figure~\ref{f6}b, if the cosine similarity exceeds a predefined threshold $\tau$, the new entity is merged into the most similar existing entity. This merging involves appending the section ID to the attribute of the existing entity. Otherwise, the new entity is added to the local schema as a distinct entry, complete with a unique label, element identifier, and source section attributes.

As illustrated in Figure~\ref{f6}c, each new entity refined by the iEntities Refiner is stored in the local schema, either as a head entity $\mathbf{h}$ or a tail entity $\mathbf{t}$. \revc{Thus,}
\begin{equation}
E = \left\{\mathbf{h}_{a}\in\mathbb{R}^{d}\mid a=1,\ldots,p\right\}
\cup
\left\{\mathbf{t}_{b}\in\mathbb{R}^{d}\mid b=1,\ldots,q\right\},~\text{where } p+q=N.
\label{e2}
\end{equation}

Finally, the embeddings for triples are created. See Figure~\ref{f6}c. Embeddings representing the head entity $\mathbf{h}_1$, relation $\mathbf{r}_1$, and tail entity $\mathbf{t}_1$ are concatenated horizontally to create the triple embeddings $\mathbf{v}_1$:
\begin{equation}
\mathbf{v}_{1}=[\mathbf{h}_{1},\,\mathbf{r}_{1},\,\mathbf{t}_{1}],~ \mathbf{v}_{1}\in\mathbb{R}^{1\times 3d}.
\label{e3}
\end{equation}

These embeddings are exported from Neo4j and stored in a SQLite database to facilitate the downstream question-answering systems.

\subsubsection{Documents Navigator Graph}\label{s424}
The Document Navigator Graph (DNG) is designed to efficiently locate and retrieve specific items from OSHA standards by indexing section IDs within a graph database. This graph captures two explicit types of regulatory relationships, as detailed in Section~\ref{bg}: (1) Cross-reference relationships linking different provisions, and (2) Parent-child hierarchical structures aligning with OSHA's standard naming and numbering conventions. The relationships are predefined (e.g., has, refers\_to)

\begin{figure}[h]
\centering
\includegraphics[width=\columnwidth]{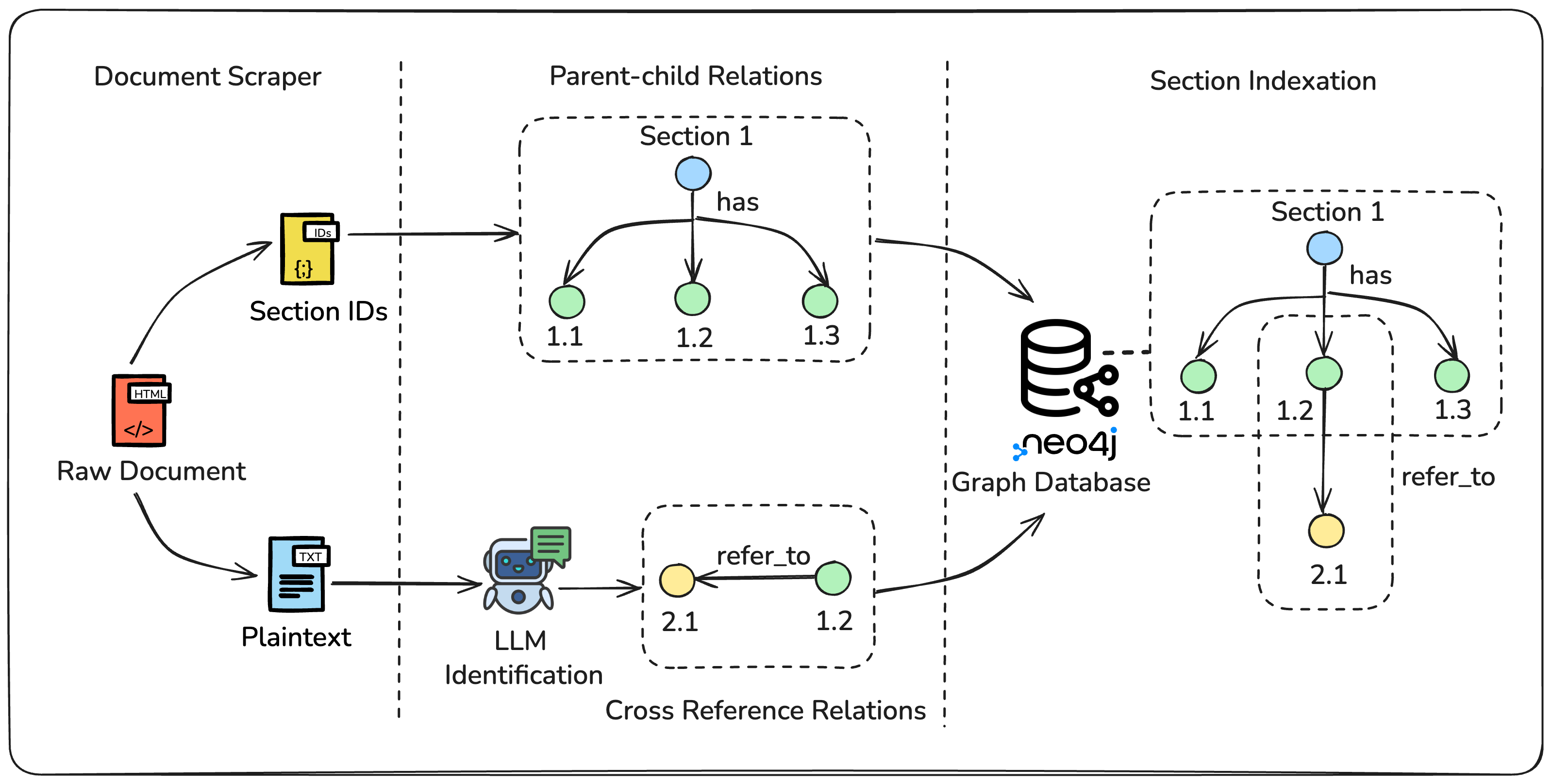}
\caption{Construction of Document Navigator Graph}
\label{f7}
\end{figure}

As illustrated in Figure~\ref{f7}, upon extraction by the Document Scraper, parent-child relationships are directly stored in the Neo4j Graph DBMS using the relationship label `has'. Conversely, identifying cross-reference relationships is integrated with the entity extraction process described in Entity Extraction. During this process, the LLM (Figure \ref{f5}) not only extracts entities but also identifies all referenced provisions within the input text. Given the variability in textual expressions of these references, the prompt for the LLM includes in-context few-shot instructions with a comprehensive set of reference patterns to ensure accurate extraction. To further ensure consistency, a rule-based post-extraction validation (Figure \ref{f5}) process ensures that every section ID conforms to the official CFR format and that all relationships adhere to the predefined schema (e.g., has, refers\_to). This ``schema-locked'' approach prevents hallucinated nodes or relations. These identified relationships are subsequently recorded in the Neo4j Graph DBMS. This multi-layered indexing strategy enables efficient data retrieval, enhancing the effectiveness of subsequent graph traversal queries and citation discovery tasks.

\subsection{Hybrid Retrieval Mechanism}
To present users with accurate, concise, and contextually appropriate information, a hybrid retrieval mechanism augmented with dual knowledge graphs is proposed to enable an incorporated QA system. In this mechanism, an LLM agent is employed to extract major entities and triples from user queries, employing the same prompt configurations as the entity extraction agent and relationship extraction agent (Section \ref{s422}), respectively. In addition, a swapping mechanism is introduced to increase the system's matching robustness to the diverse voice or sentence structure of user queries. For example, the equivalent semantic meaning expressed in the active voice (e.g., ``A competent person can design the ramp.'') or in the passive voice (e.g., ``The ramp can be designed by a competent person.'') may produce subject-object inversions in triples. To address this, the system generates symmetric triples by swapping the source and target entities.

Both entities and triples extracted from the user query are mapped to the most relevant elements in the ENG by computing similarity scores. The process begins by calculating cosine similarity between target entities and those stored in the local schema. Top five entities with the similarity score greater than 0.5 are retrieved, and their section IDs are collected into a candidate set. These sections capture provisions with implicit connections to each other and the query due to shared terminology (entity). Due to prior entity merging by the iEntities Refiner, one entity may be associated with multiple section IDs, potentially leading to an overly broad set of relevant sections. To narrow this down, a second round of cosine similarity comparison was conducted on triples. Similarly, the top five triples with the highest similarity scores (greater than 0.5) are selected, and their corresponding section IDs are consolidated into another candidate set. This parallel retrieval refines entity similarity results because shared terminology often consists of common terms appearing across multiple provisions, whereas triples focus the search more precisely. Additionally, triple similarity search identifies related regulatory scope: a triple such as (``excavation,'' ``subject\_to,'' ``heavy rain'') may surface complementary safety provisions governing adverse weather conditions. Only the intersection of two candidate sets would be considered, ensuring highly relevant provisions with implicit relationships are included.

The selected section IDs are crucial, as the section texts retrieved according to them will serve as entry points for multi-hop indexing. Each retrieved section is examined to determine whether it references other sections. All the connected sections are added to the pool and iteratively examined until exhausted. This process is powered by the established DNG, using cypher queries \texttt{MATCH (a)-[:refers\_to]-$>$(b) WHERE a.name = \$section\_name RETURN b.name AS name} to identify connections. All matched sections are included as candidate inputs to the LLM agent, and the process continues until all relevant sections are processed.

Finally, an LLM agent processes all retrieved texts by selectively retaining only those segments that are directly relevant to the user's query, and generate a coherent and focused summary response. To ensure both the relevance and accuracy of the answer, the model is instructed to first review all cited sources and filter out any references deemed irrelevant or unhelpful. It then synthesizes a comprehensive response using only the pertinent references, maintaining clarity, logical structure, and alignment with the original question. Throughout this process, the original texts and their citations are preserved to support transparency and facilitate traceability.

\section{QA Implementation}
\revc{This section presents the implementation of the proposed system, in which the constructed graphs and the LLM are integrated into an end-to-end pipeline for multi-hop QA.}

\subsection{Interface of QA System}
\begin{figure}[h]
\centering
\includegraphics[width=\columnwidth]{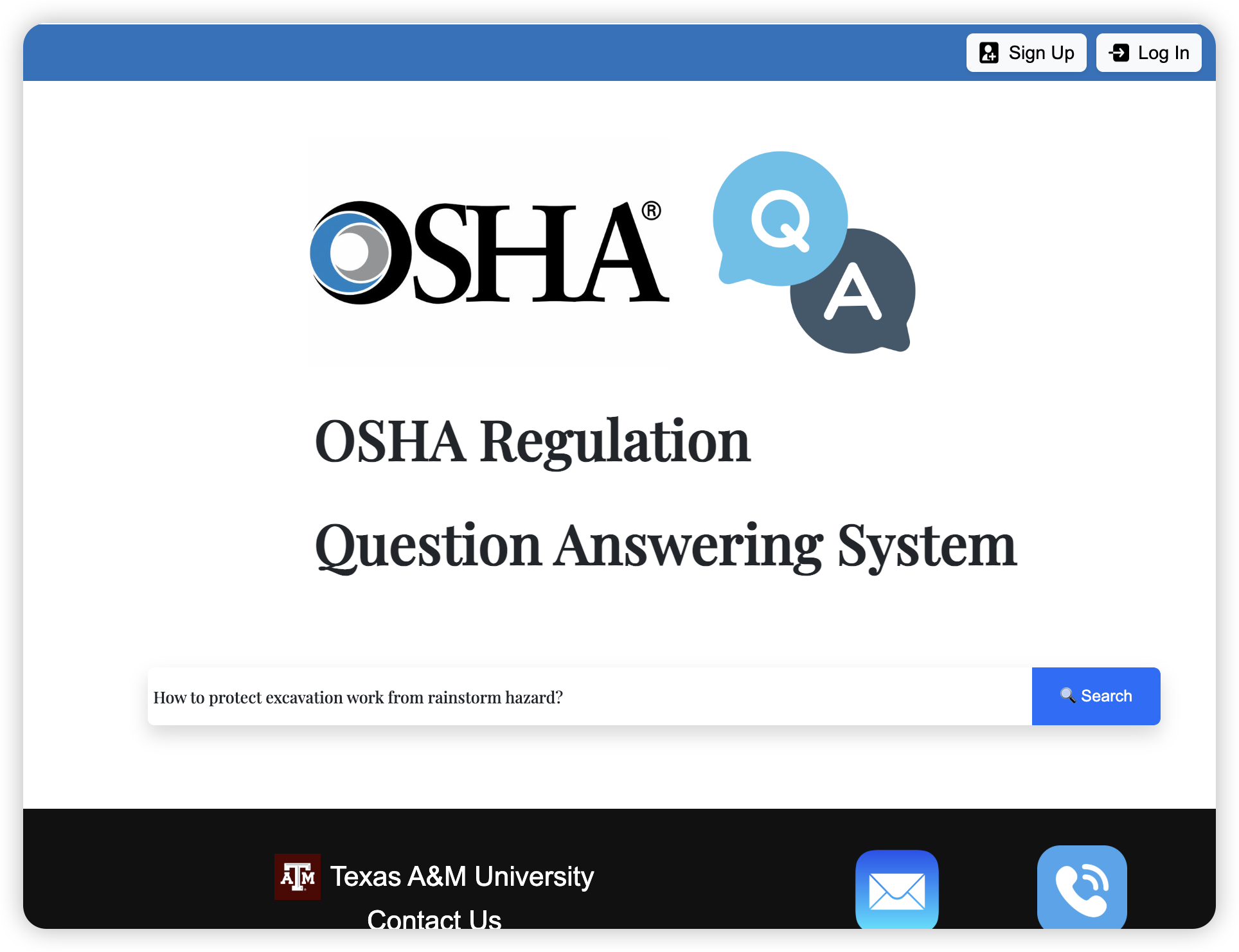}
\caption{Designed Interface of the OSHA Regulation QA System}
\label{f9}
\end{figure}

\begin{figure}[h]
\centering
\includegraphics[width=\columnwidth]{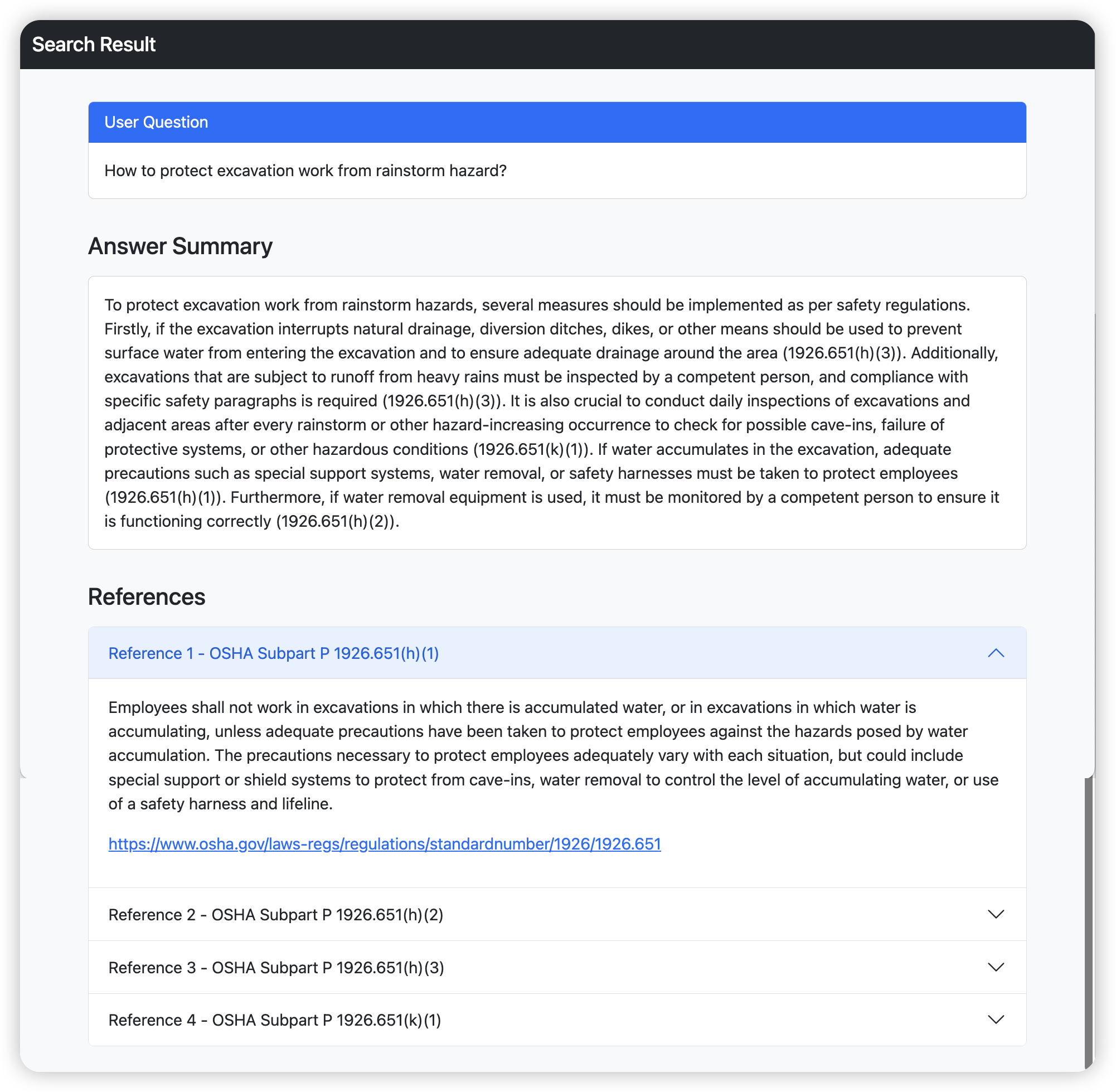}
\caption{Search Result for a Sample User Query}
\label{f10}
\end{figure}

Figure~\ref{f9} presents the user interface of the QA system, which comprises a registration bar, a search box, and a footer containing contact information. Users enter their OSHA-related queries in the search box and click the ``Search'' button, after which they are redirected to a results page. An example of this page is shown in Figure~\ref{f10}. Following the summary, the system provides references that include the original regulation text, the corresponding section IDs, and direct links to these materials.

\subsection{Case Study}
\begin{figure*}
\centering
\includegraphics[width=0.85\textwidth]{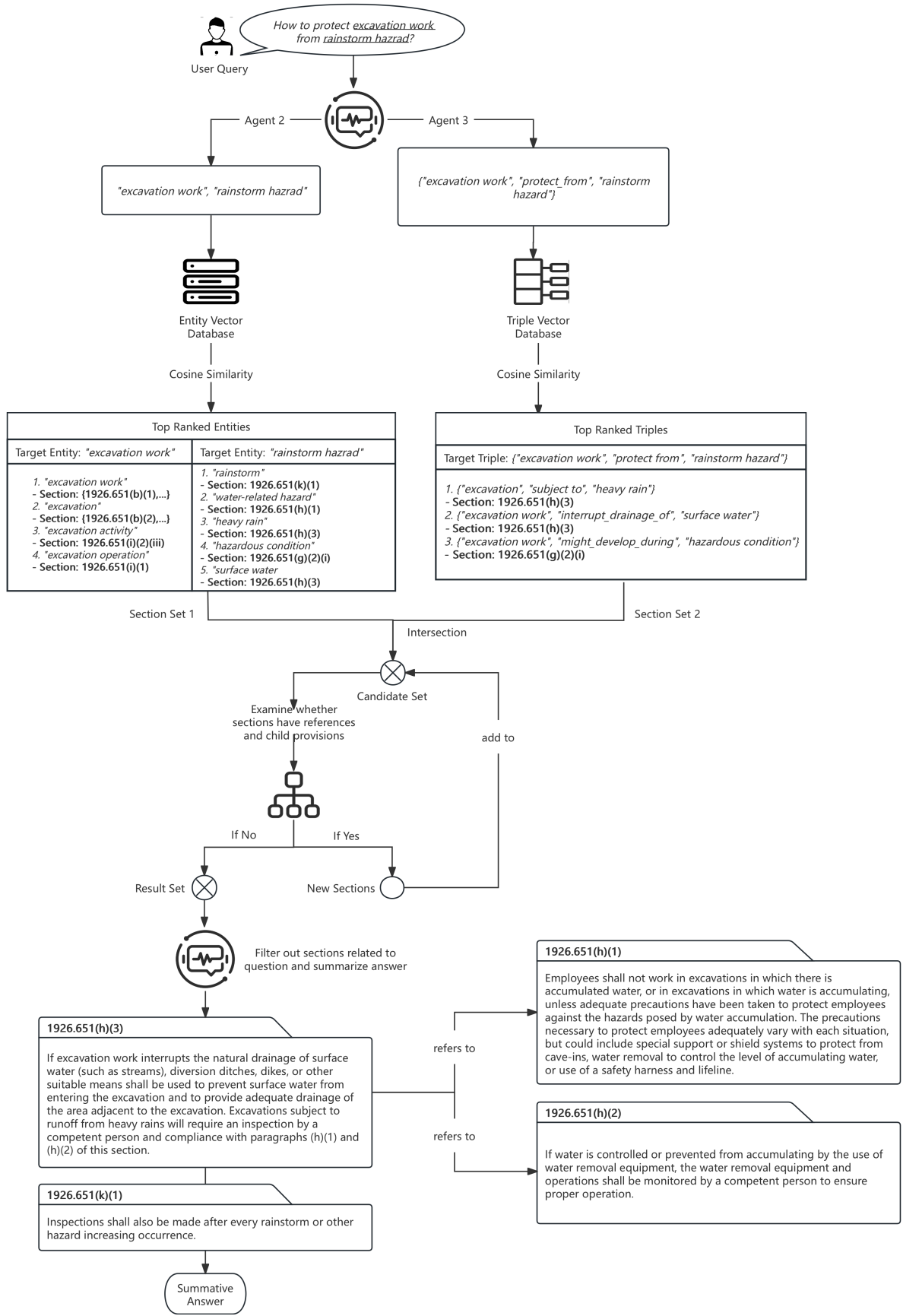}
\caption{Query Processing Pipeline of QA System}
\label{f8}
\end{figure*}

To illustrate the effectiveness of \revc{the} proposed approach, \revc{a representative case study has been conducted Figure~\ref{f8}}. Consider the user query: ``How to protect excavation work from rainstorm hazard?'' This query contains two key entities, ``excavation work'' and ``rainstorm hazard,'' connected by the relation ``protect\_from,'' together forming a complete semantic triple.

In \revc{the} local entity similarity analysis, four entities exhibited the highest semantic similarity to ``excavation work'': (1) ``excavation work'' (similarity = 1.000), (2) ``excavation'' (0.8977), (3) ``excavation activity'' (0.8977), and (4) ``excavation operation'' (0.8859). Similarly, the five most similar entities to ``rainstorm hazard'' were identified as: (1) ``rainstorm'' (0.7506), (2) ``water-related hazard'' (0.6626), (3) ``heavy rain'' (0.6139), (4) ``hazardous condition'' (0.5742), and (5) ``surface water'' (0.5731). Based on these entity pairs, several relevant triples were retrieved from the Entity Network Graph, including: (1) (``excavation'', ``subject\_to'', ``heavy rain'') (similarity = 0.6343), (2) (``excavation work'', ``interrupt\_drainage\_of'', ``surface water'') (0.6145), and (3) (``excavation work'', ``might\_develop\_during'', ``hazardous condition'') (0.5835).

Next, \revc{the Entity Network Graph was queried} to determine the source sections associated with each of these entities and triples. By intersecting the corresponding section sets, \revc{a refined candidate set of sections was obtained}. To ensure comprehensiveness, \revc{each section in the candidate set was further examined} using the Document Navigator Graph to check for cross-references or child provisions. For instance, the query led directly to Sections 1926.651(h)(3) and 1926.651(k)(1). However, Section 1926.651(h)(3) explicitly references Sections 1926.651(h)(1) and 1926.651(h)(2), which collectively form a complete regulatory chain. As such, all three sections were included for further processing. This recursive exploration continues until all potentially relevant provisions are identified.

Finally, the full text of all retrieved provisions, along with their source locations, is passed to the GPT model. The model then filters out irrelevant content and generates a summative answer that directly addresses the original user query.

\section{Experiment}
\revc{This section describes the experimental setup used to evaluate the proposed system, in which the dataset, baselines, and evaluation metrics are specified.}
\subsection{Experiment Design}
Using GPT-4o as the backbone of the LLM agents, \revc{BifrostRAG was evaluated} against three state-of-the-art solutions: LLM-only (without RAG), vector-based RAG using OpenAI embeddings, and graph-based RAG implemented with Neo4j. GPT-4o without RAG was selected as the baseline to evaluate the capabilities of a state-of-the-art LLM based solely on its pre-trained knowledge, establishing a fundamental lower-bound performance benchmark for \revc{the} evaluation. Although prior research has demonstrated GPT-4o's proficiency in close-ended, standardized safety assessments \citep{sammour2024responsible}, its effectiveness in open-ended questions—particularly those requiring interpretation of complex safety regulations—remains underexplored. \revc{OpenAI's vector-based RAG and Neo4j's graph-based RAG implementations were selected} as they represent widely adopted, cutting-edge solutions for building vector-based and graph-based RAG systems, respectively. Also, they are off-the-shelf applications accessible for researchers and practitioners, ensuring \revc{the} evaluation results provide immediately practical insights for real-world applications. Within the domain of graph-based RAG, Neo4j's pipeline emerges as the most functionally aligned baseline for \revc{the} task. \revc{While alternative solutions such as Microsoft’s GraphRAG are considered, it is found that it builds a topic-based summary graph rather than an explicit entity–relation knowledge graph}; its nodes represent thematic communities and summaries, and retrieval is organized around these topic clusters and their condensed descriptions. Moreover, \revc{the} proposed Bifrost RAG system builds upon Neo4j’s open-source graph database infrastructure. This alignment was intentional, as it \revc{allowed for isolated and rigorous assessment of the impact of the core contributions}: the dual-graph architecture and the specialized retrieval strategy, rather than to disparities in backend technology.

\revc{Vector-based RAG was configured} through OpenAI Playground, a web-based developer platform that enables customization of and interaction with OpenAI's LLMs. The platform automatically embedded uploaded documents into a vector store using OpenAI's default text-embedding-3-small model, allowing LLMs to retrieve relevant content and integrate it into responses. For the graph-based RAG implementation, \revc{documents were uploaded} to Neo4j to construct a knowledge graph. Neo4j automatically generates the knowledge graph from uploaded documents by integrating LangChain LLMGraphTransformer (version 0.4), which employed built-in general-purpose prompts to guide the LLM in identifying entities and their semantic relationships. The resulting knowledge graph was stored in Neo4j's platform. For QA, Neo4j utilizes LangChain's GraphCypherQAChain (version 0.4) to direct LLMs to first generate Cypher queries and execute them over the knowledge graph. The graph traversal process begins by matching entities to determine the appropriate subgraph for traversal. \revc{To ensure a fair comparison, GPT-4o-2024-08-06 was used as both the extraction and QA agent across all configurations. The model temperature was standardized at 0.2, and the maximum token limit was set to 1,500.} An exception was Neo4j, where full alignment with these parameters was not feasible due to its built-in configuration constraints, which may have influenced the generated outputs.

\revc{The} validation study examined four subparts OSHA 1926, as shown in Table~\ref{t6}. These subparts were chosen for three reasons. First, they address fall protection—the leading cause of construction fatalities—making them highly relevant to industry safety priorities. Second, these subparts exhibit the full spectrum of complex structural relationships found in safety regulations, creating an ideal testing environment for multi-hop QA. For example, Subpart M (Fall Protection) establishes horizontal regulations governing fall protection systems under general conditions, while Subparts L (Scaffolds), P (Excavations), and X (Stairways and Ladders) provide vertical regulations with specialized fall requirements. Third, document sizes of these subparts align with those typically used in similar studies \citep{wang2023deep,wang2023bdeep}.

To systematically evaluate model performance, \revc{a set of 93 questions was designed}. All questions in \revc{the} dataset are, by design, multi-hop. This was an intentional choice, as addressing complex, multi-hop reasoning is the core focus of \revc{this paper}. Since prior research shows that single-hop questions are largely solved by existing methods, \revc{the evaluation focuses} on the challenging queries that are most representative of the problems faced by safety professionals. 

 The specific number of questions for each subpart is indicated in Table~\ref{t4}. Table~\ref{t5} presents the distribution of multi-hop mechanisms in the set of questions. Table~\ref{t6} shows the distribution of the relationship types in the set of questions. Collectively, the questions draw relatively evenly from multiple regulatory subparts, reflecting various safety regulatory topics. Questions cover diverse relationship types manifested through all four mechanisms, aligning with the real-world reasoning patterns required in regulatory compliance tasks and providing a robust test of RAG performance on multi-hop questions.



\begin{table}[h]
\centering
\footnotesize
\caption{Selected OSHA Subparts}
\label{t4}
\begin{tabularx}{\columnwidth}{
>{\raggedright\arraybackslash}p{2cm}
>{\raggedright\arraybackslash}X
>{\centering\arraybackslash}p{0.8cm}
}
\toprule
\textbf{Subpart} & \textbf{Topic} & \textbf{Number} \\
\midrule
1926 Subpart L &
Scaffolds (1926.451, 1926.452, 1926.453, 1926.454) &
30 \\
1926 Subpart M &
Fall Protection (1926.500, 1926.501, 1926.502, 1926.503) &
27 \\
1926 Subpart P &
Excavations (1926.651, 1926.652) &
17 \\
1926 Subpart X &
Stairways and Ladders (1926.1051, 1926.1052, 1926.1053) &
19 \\
\midrule
\textbf{Total} & & \textbf{93} \\
\bottomrule
\end{tabularx}
\end{table}

\begin{table}[h]\linespread{1.05}\selectfont
\centering
\footnotesize
\caption{Distribution of Multi-hop Mechanisms in the Question Set}
\label{t5}
\begin{tabular}{p{2.1cm}p{2.5cm}p{0.8cm}p{1.2cm}}
\toprule
\textbf{Multi-hop Type} & \textbf{Mechanism} &
\textbf{Number} &
\textbf{Percentage} \\
\midrule
Implicit Multi-hop  & Cross-reference & 21 & 22.58\% \\
 & Hierarchical structure & 31 & 33.33\% \\
Explicit Multi-hop & Shared terminology & 26 & 27.96\% \\
 & Relevant scop & 31 & 33.33\% \\
\bottomrule
\end{tabular}\\
\vspace{0.5em}
\begin{minipage}{\linewidth}
\scriptsize \raggedright
Note: The percentages sum to more than 100\% since some complex questions may involve multiple mechanisms simultaneously.
\end{minipage}
\end{table}

\begin{table}[h]\linespread{1.05}\selectfont
\centering
\footnotesize
\caption{Distribution of Relationship Types in the Question Set}
\label{t6}
\begin{tabular}{p{3cm}p{1.5cm}p{2.5cm}}
\toprule
\textbf{Relationship Type} & \textbf{Number} & \textbf{Percentage} \\
\midrule
Exception & 17 & 18.28\% \\
Conditional & 29 & 31.18\% \\
Complementary & 23 & 24.73\% \\
Cross-Cutting & 28 & 30.11\% \\
\bottomrule
\end{tabular}\\
\vspace{0.5em}
\begin{minipage}{\linewidth}
\scriptsize \centering
Note: The percentages sum to more than 100\% as some questions involve more than one type of relationship.
\end{minipage}

\end{table}

\revc{Ten open-ended questions were first drawn from} authoritative OSHA resources, each accompanied by answers citing multiple sections of OSHA 1926. Using these questions as templates, \revc{the remaining 83 questions were developed} in collaboration with a construction safety expert. The expert is an OSHA-authorized trainer and a Certified Safety Professional (CSP) with over eight years of industry experience. As each question was formulated, all relevant section IDs were annotated to establish a ground-truth reference. To ensure accuracy, \revc{a two-step cross-check was performed} within the research team. First, \revc{the annotated ground-truth section IDs were verified} for correctness. Second, \revc{additional sections were systematically reviewed} to identify any potentially overlooked relevant references. Through expert deliberation and iterative refinement, a set of predefined ground-truth answers was established for performance evaluation.

\subsection{Evaluation Metrics}
In the experiment, each model was instructed to generate answers with corresponding reference section IDs. \revc{Model performance in answering multi-hop QA was evaluated} using three objective metrics: (1) Precision rate: The proportion of retrieved section IDs that are correct; (2) Recall rate: the proportion of correct section IDs that were successfully retrieved; and (3) F1 score: the harmonic mean of precision and recall, representing a balanced measure of retrieval accuracy. Since all tested models use the GPT-4o model, the primary differentiator lies in each RAG system's information retrieval capability. When provided with accurate references, GPT-4o consistently provided accurate written answers. \revc{This} evaluation framework therefore focuses on capturing whether systems can comprehensively identify relevant provisions without introducing irrelevant information that could mislead compliance assessments.

\begin{equation}
Precision = \frac{|Correct\ Sections\ in\ Answer|}{|Total\ Sections\ in\ Answer|}
\label{e4}
\end{equation}

\begin{equation}
Recall = \frac{|Correct\ Sections\ in\ Answer|}{|Total\ Required\ Sections|}
\label{e5}
\end{equation}

\begin{equation}
F1 = 2\cdot\frac{Recall\cdot Precision}{Recall + Precision}
\label{e6}
\end{equation}

\section{Experimental Results}
\revc{This section reports the experimental results and performance analyses, including the validation of both the dual-graph construction pipeline and the QA outcomes. The proposed system is further examined through an ablation study to observe the system’s behavior under different settings.}
\subsection{Dual Knowledge Graph Construction}
\textbf{Validation:} To evaluate the automated knowledge-graph construction pipeline, \revc{experiments were conducted} on OSHA 1926 Subparts L, M, P, and X. For each subpart, 20 provisions were randomly sampled. Domain experts created gold-standard annotations under a unified extraction schema. Using the same instructions and schema, the automated system (GPT-4o) produced machine annotations. For each subpart, \revc{precision, recall, and F1 were computed} for entity extraction and triple creation. Results are reported in Table~\ref{t7} and Table~\ref{t8}. Precision, recall, and F1 scores were micro-averaged within each subpart and macro-averaged across subparts. As expected, triple creation proved marginally more challenging than entity extraction, yielding on average 1.5–2.8 percentage points lower precision, 1.2–3.6 points lower recall, and 1.1–3.1 points lower F1. This performance gap is anticipated, as triple generation is inherently dependent on the preceding accuracy of entity extraction.


\begin{table*}[t]
\linespread{1.05}\selectfont
\centering
\footnotesize
\caption{Performance of Machine Annotated Entity Across Subparts}
\label{t7}
\begin{tabular}{p{3cm}p{3cm}p{2cm}p{2cm}p{2cm}}
\toprule
\textbf{Subpart} & \textbf{Number of Provisions} & \textbf{Precision} & \textbf{Recall} & \textbf{F1 Score} \\
\midrule
1926 Subpart L & 20 & 94.07\% & 92.30\% & 92.49\% \\
1926 Subpart M & 20 & 95.83\% & 94.79\% & 95.19\% \\
1926 Subpart P & 20 & 94.69\% & 92.16\% & 93.08\% \\
1926 Subpart X & 20 & 95.37\% & 93.52\% & 94.36\% \\
Macro Avg. & 80 & 94.99\% & 93.19\% & 93.78\% \\
\bottomrule
\end{tabular}
\end{table*}

\begin{table*}[t]\linespread{1.05}\selectfont
\centering
\footnotesize
\caption{Performance of Machine Annotated Triple Across Subparts}
\label{t8}
\begin{tabular}{p{3cm}p{3cm}p{2cm}p{2cm}p{2cm}}
\toprule
\textbf{Subpart} & \textbf{Number of provisions } & \textbf{Precision} & \textbf{Recall} & \textbf{F1 Score} \\
\midrule
1926 Subpart L & 20 & 92.60\% & 91.08\% & 91.38\% \\
1926 Subpart M & 20 & 93.75\% & 91.15\% & 92.07\% \\
1926 Subpart P & 20 & 92.29\% & 90.40\% & 90.53\% \\
1926 Subpart X & 20 & 92.57\% & 90.99\% & 91.69\% \\
Macro Avg. & 80 & 92.80\% & 90.91\% & 91.42\% \\
\bottomrule
\end{tabular}
\end{table*}

To assess the stability and reliability of the machine annotation pipeline, \revc{95\% confidence intervals (CIs) were computed} for precision, recall, and F1 across all subparts. For both entity extraction and triple creation tasks, the 95\% CI widths remained within 4 percentage points for all metrics. Specifically, the maximum CI width was 3.79\% for triple-level precision, whereas the minimum was 3.17\% for entity-level recall.

These tight 95\% CIs, together with the high absolute scores (all exceeding 90\%), suggest that the machine annotation framework achieves statistically robust and reliable performance. This level of stability supports its applicability to automated information extraction for knowledge graph construction.

\begin{figure}[h]
\centering
\includegraphics[width=\columnwidth]{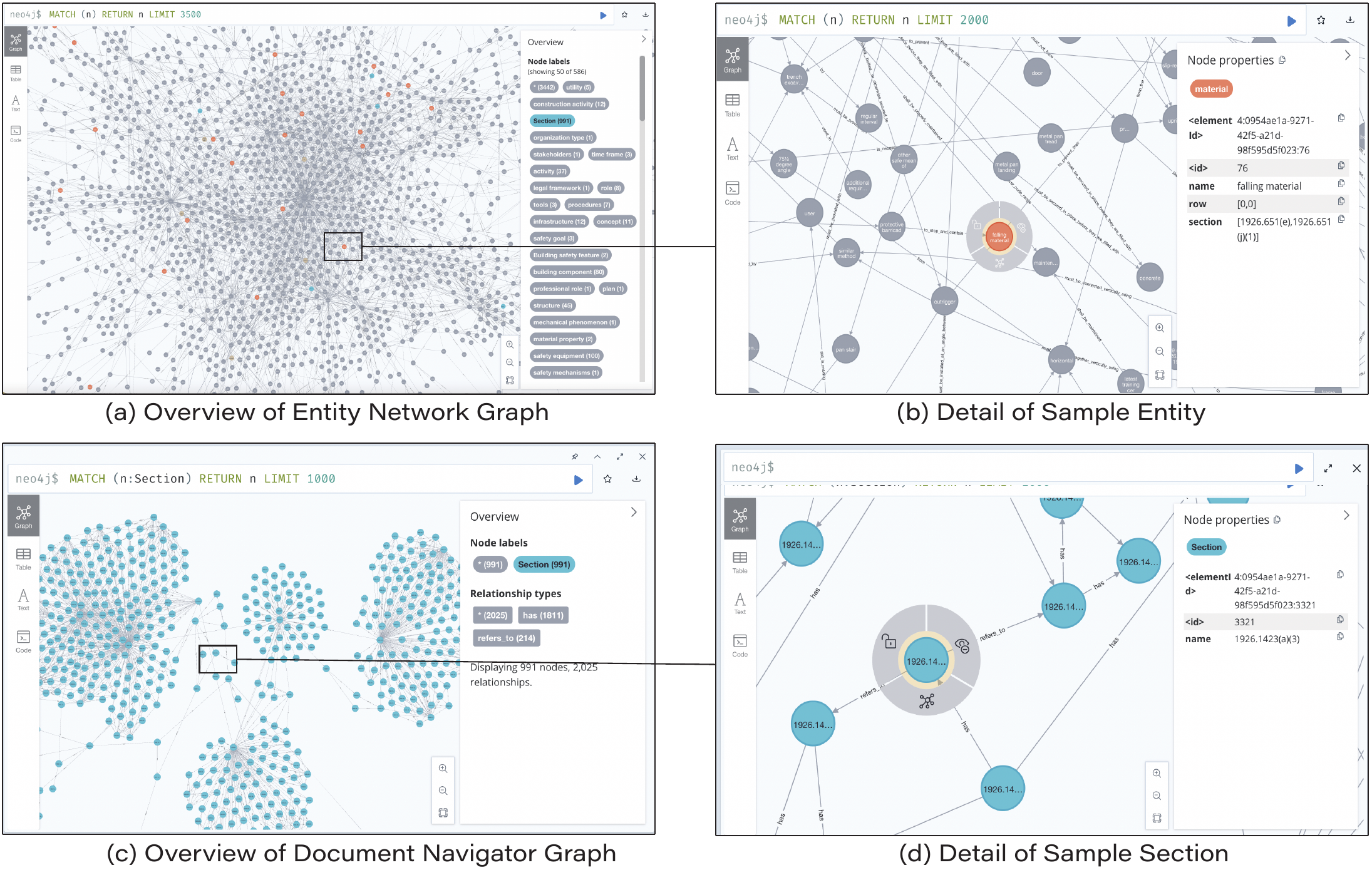}
\caption{Visualization of the Generated Dual Knowledge Graphs in Neo4j}
\label{f11}
\end{figure}

\textbf{Graph Database:} After validating the machine annotation pipeline, \revc{it was applied to} large-scale knowledge graph construction across the complete OSHA 1926 Subparts L, M, P, and X. This implementation resulted in two knowledge graphs: the Document Navigator Graph and the Entity Network Graph. The Document Navigator Graph is a hierarchical structure organizing all 991 section indexes (represented as blue nodes in Figure~\ref{f11}), and 2025 regulatory relationships. It uses two relationship types to connect sections: parent-child, and cross-reference. This knowledge graph captures relationships explicitly embedded in the regulation for retrieval during QA. The Entity Network Graph operates at the phrase level, extracting entities, such as ``falling objects'' and ``barricade,'' and defining their interrelationships. \revc{The knowledge graph constructed} in Neo4j consists of 3,442 nodes and 6,221 relationships, as shown in Figure~\ref{f11}. This network captures implicit connections between sections, even across subparts, that are not explicitly stated through parent-child, or cross-reference relationships but are relevant due to shared entities and overlapping scopes. The Document Navigator Graph and the Entity Network Graph are linked through the section number, which serves as a shared property of nodes in both graphs.

\subsection{QA System Performance Comparison}
\begin{figure}[h]
\centering
\includegraphics[width=\columnwidth]{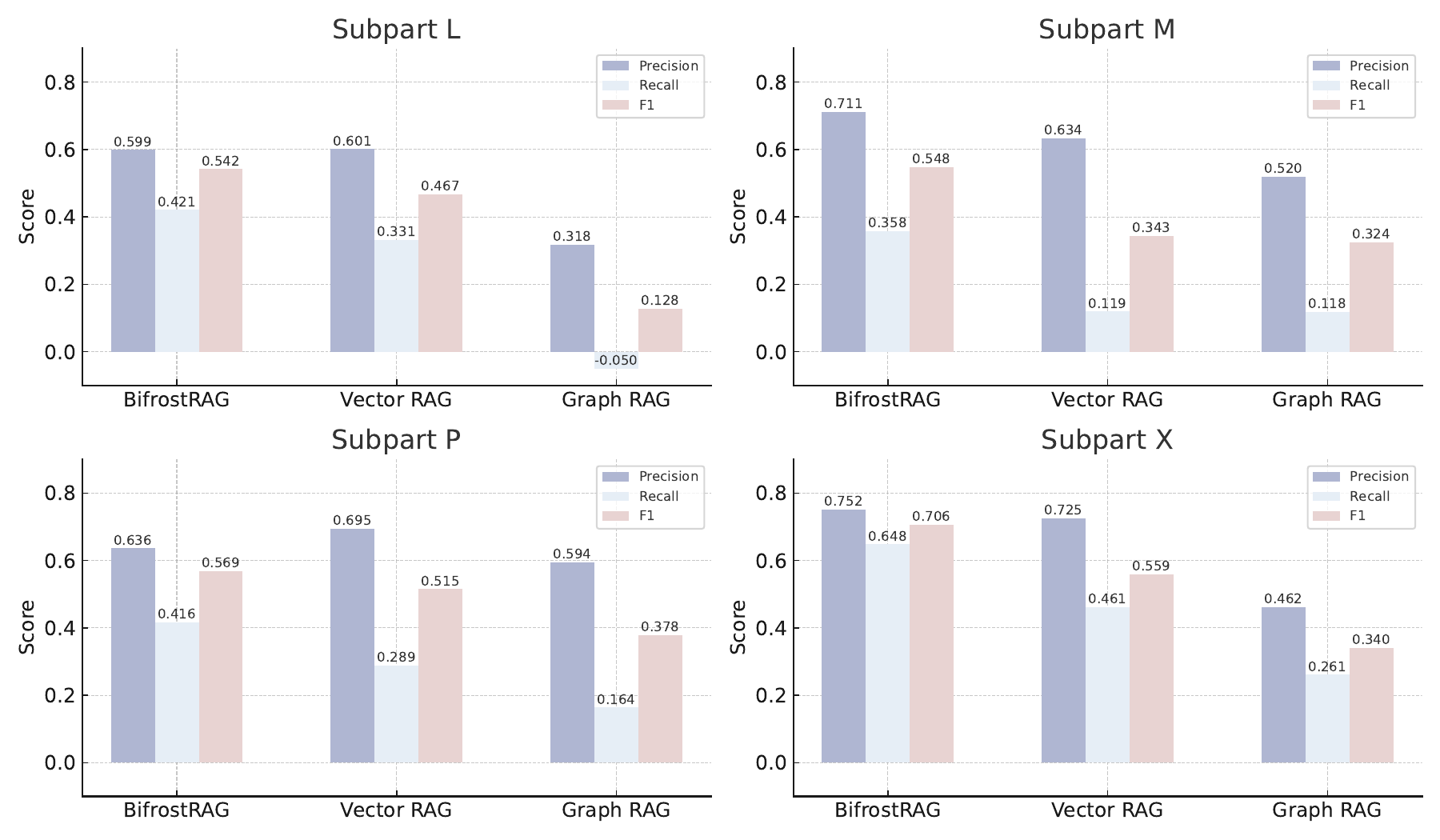}
\caption{Performance Comparison Across OSHA Subparts}
\label{f12}
\end{figure}

\revc{To demonstrate functionality and validate performance of proposed QA system, the same question set was run on BifrostRAG and three comparative models.} Table~\ref{t9} presents the descriptive statistics for precision, recall rates, and F1 score, along with their standard deviations, across four models for each subpart and overall performance. The results indicate that BifrostRAG outperforms vector-based RAG, Neo4j, and a naive implementation of GPT-4o. This improvement is confirmed through statistical analysis. The bootstrap confidence intervals in Table~\ref{newt10} indicate that the overall performance estimates are already quite stable, with the 95\% CIs for BifrostRAG's macro-averaged precision, recall, and F1 all lying within $\pm5\%$. Given the non-normality of \revc{the} data, \revc{the Friedman non-parametric ANOVA test is employed}, followed by post-hoc pairwise analysis with Bonferroni corrections. The results are shown in Table~\ref{t11} and Table~\ref{t12}. In addition to p-values, \revc{adjusted p-values are reported} to mitigate the risk of Type I errors resulting from multiple comparisons. Figure~\ref{f12} illustrates the performance improvements of the three tested RAG solutions compared to the baseline approach (GPT-4o without RAG) across four OSHA subparts.

\begin{table*}[t]\linespread{1.05}\selectfont
\centering
\footnotesize
\setlength{\tabcolsep}{1pt}
\renewcommand{\arraystretch}{1.1}
\linespread{1.1}
\selectfont
\caption{Results of Evaluation Metrics for Compared Models}
\label{t9}
\begin{tabular}{l
  >{\centering\arraybackslash}c
  >{\centering\arraybackslash}p{2.4cm}
  >{\centering\arraybackslash}p{2.4cm}
  >{\centering\arraybackslash}p{2.4cm}
  >{\centering\arraybackslash}p{2.4cm}}
\toprule
\textbf{Subpart} & \textbf{Metrics} & \textbf{BifrostRAG} & \textbf{\shortstack{Vector RAG\\(OpenAI)}} & \textbf{\shortstack{Graph RAG\\(Neo4j)}}
 & \textbf{No RAG} \\
\midrule
Subpart L & P & 87.0\% (0.327) &  \textbf{87.2\% (0.248)} &  58.9\% (0.475) &  27.1\% (0.277) \\
Fall Protection & R & \textbf{ 84.0\% (0.349)} &  75.0\% (0.325) &  36.9\% (0.378) &  41.9\% (0.429) \\
 & F1 & \textbf{ 84.1\% (0.338)} &  76.6\% (0.276) &  42.7\% (0.388) &  29.9\% (0.287) \\
 \addlinespace[3pt]
Subpart M & P & \textbf{ 93.2\% (0.245)} &  85.5\% (0.326) &  74.1\% (0.361) &  22.1\% (0.215) \\
Scaffolding & R & \textbf{ 75.9\% (0.298)} &  52.1\% (0.294) &  52.0\% (0.303) &  40.1\% (0.397) \\
 & F1 & \textbf{ 81.6\% (0.263)} &  61.1\% (0.277) &  59.3\% (0.305) &  26.8\% (0.250) \\
 \addlinespace[3pt]
Subpart P & P & 94.1\% (0.166) & \textbf{100\% (0.000)} &  89.8\% (0.192) &  30.5\% (0.183) \\
Excavation & R & \textbf{ 95.1\% (0.141) }&  82.4\% (0.246) &  69.8\% (0.301) &  53.5\% (0.383) \\
 & F1 & \textbf{  93.6\% (0.145)} &  88.1\% (0.179) &  74.4\% (0.239) &  36.6\% (0.225) \\
 \addlinespace[3pt]
Subpart X & P & \textbf{ 100\% (0.000)} &  97.4\% (0.115) &  71.1\% (0.451) &  24.8\% (0.345) \\
Stairways and Ladders & R & \textbf{ 92.8\% (0.189)} &  74.1\% (0.276) &  54.1\% (0.404) &  28.0\% (0.353) \\
 & F1 & \textbf{ 95.0\% (0.135)} &  80.3\% (0.200) &  58.5\% (0.397) &  24.4\% (0.310) \\
 \addlinespace[3pt]
Overall & P & \textbf{ 92.8\% (0.240)} &  91.1\% (0.236) &  71.4\% (0.407) &  25.8\% (0.259) \\
 & R & \textbf{  85.5\% (0.282)} &  69.5\% (0.311) &  50.8\% (0.363) &  40.7\% (0.399) \\
 & F1 & \textbf{ 87.3\% (0.257)} &  75.0\% (0.262) &  56.5\% (0.356) &  29.1\% (0.270) \\
\bottomrule
\end{tabular}
\end{table*}


\begin{table}[t]
\centering
\footnotesize
\setlength{\tabcolsep}{4pt}
\renewcommand{\arraystretch}{1.05}
\caption{Overall performance of the QA systems with 95\% bootstrap confidence intervals}
\label{newt10}
\begin{tabularx}{\columnwidth}{l >{\centering\arraybackslash}X >{\centering\arraybackslash}X >{\centering\arraybackslash}X}
\toprule
 & \textbf{BifrostRAG} & \textbf{Vector-based RAG} & \textbf{Graph-based RAG} \\
\midrule
Precision & 92.8$\pm$3.6\% & 91.1$\pm$6.9\% & 71.4$\pm$10.3\% \\
Recall    & 85.5$\pm$4.5\% & 69.5$\pm$8.2\% & 50.8$\pm$8.7\% \\
F1-score  & 87.3$\pm$3.9\% & 75.0$\pm$7.2\% & 56.5$\pm$8.5\% \\
\bottomrule
\end{tabularx}
\end{table}

\begin{table}[t]\linespread{1.05}\selectfont
\centering
\footnotesize
\setlength{\tabcolsep}{10pt}
\renewcommand{\arraystretch}{1.05}
\linespread{1.1}
\selectfont
\caption{Results of Friedman Test}
\label{t11}
\begin{tabular}{lccc}
\toprule
\textbf{Metric} & \textbf{Chi-Square} & \textbf{Asymp.Sig} & \textbf{Exact. Sig} \\
\midrule
Precision & 25.597 & $<$0.001*** & $<$0.001*** \\
Recall & 47.528 & $<$0.001*** & $<$0.001*** \\
F1 score & 57.475 & $<$0.001*** & $<$0.001*** \\
\bottomrule
\end{tabular}\\
\vspace{0.5em}
\begin{minipage}{\linewidth}
\scriptsize \centering
Note: * indicates p $<$ 0.05; ** indicates p $<$ 0.005; *** indicates p $<$ 0.001
\end{minipage}
\vspace{-3em}
\end{table}

\begin{table*}[t]\linespread{1.05}\selectfont
\centering
\footnotesize
\renewcommand{\arraystretch}{1.1}
\linespread{1.1}
\selectfont
\caption{Result of Post-hoc Analysis}
\label{t12}
\begin{tabular}{llcccccc}
\toprule
\textbf{Metric} & \textbf{Model} &
\multicolumn{3}{c}{\textbf{Pairwise Comparison Sig}} &
\multicolumn{3}{c}{\textbf{Pairwise Comparison Adj. Sig}} \\
\cmidrule(r){3-5} \cmidrule(l){6-8}
& & BifrostRAG & Vector RAG & Graph RAG & BifrostRAG & Vector RAG & Graph RAG \\
\midrule
\textbf{Precision} & BifrostRAG & 1 &  &  & 1 &  &  \\
& Vector RAG         & 0.634 & 1 &  & 1 & 1 &  \\
& Graph RAG          & 0.003** & 0.014* & 1 & 0.010** & 0.042* & 1 \\

\textbf{Recall} & BifrostRAG & 1 &  &  & 1 &  &  \\
& Vector RAG         & 0.002** & 1 &  & 0.007** & 1 &  \\
& Graph RAG          & $<$0.001*** & 0.009** & 1 & $<$0.001*** & 0.028* & 1 \\

\textbf{F1 Score} & BifrostRAG & 1 &  &  & 1 &  &  \\
& Vector RAG         & $<$0.001*** & 1 &  & $<$0.001*** & 1 &  \\
& Graph RAG          & $<$0.001*** & 0.016* & 1 & $<$0.001*** & 0.047* & 1 \\
\bottomrule
\end{tabular}\\
\scriptsize Note: * indicates p $<$ 0.05; ** indicates p $<$ 0.005; *** indicates p $<$ 0.001\hfill
\end{table*}

BifrostRAG achieves an overall precision of 92.8\% and an overall recall of 85.5\%, surpassing vector-based RAG (OpenAI Playground) (precision = 91.1\%, recall = 69.5\%) and significantly outperforming general-purpose knowledge graph-based RAG (Neo4j) (precision = 71.4\%, recall = 50.8\%). Post-hoc analysis reveals that the recall improvement over vector-based RAG is statistically significant ($p = 0.002$**, adjusted $p = 0.007$**), while the increase in precision (+1.6\%) is marginal and not statistically significant. Compared to Neo4j, \revc{the proposed method} demonstrates a 21.4\% improvement in precision and a 34.7\% increase in recall, with post-hoc pairwise comparisons confirming both differences as statistically significant. In terms of F1-score results, BifrostRAG achieved the highest score of 87.3\% (SD = 0.257), markedly outperforming the vector-based RAG (75.0\%, SD = 0.262) and graph-based RAG (56.5\%, SD = 0.356). The subsequent Friedman test and post-hoc analysis further substantiated these differences, revealing that the F1-score distinctions between BifrostRAG and both comparison models were statistically significant (adjusted sig. $<$ 0.001).

Furthermore, empirical results suggest that BifrostRAG (SDP = 0.240; SDR = 0.282) and vector-based RAG (SDP = 0.236; SDR = 0.311) exhibit comparable stability across multiple subparts, whereas graph-based RAG (SDP = 0.407; SDR = 0.363) shows greater performance fluctuations. Notably, GPT-4o without RAG performs the worst, achieving only 25.8\% precision and 40.7\% recall. When handling multi-hop questions and generating responses with citations, GPT-4o without RAG exhibits significant hallucination issues, including fabricating section IDs, providing incomplete content, and, more concerningly, generating misinformation. Another observation emerges from Figure~\ref{f12}. While BifrostRAG, OpenAI Playground, and Neo4j generally improve upon the GPT-4o baseline across all four subparts, Neo4j unexpectedly exhibits a decline in recall (0.050) relative to the GPT-4o baseline within subpart L.

\subsection{Error Analysis}
\begin{figure}[h]
\centering
\includegraphics[width=\columnwidth]{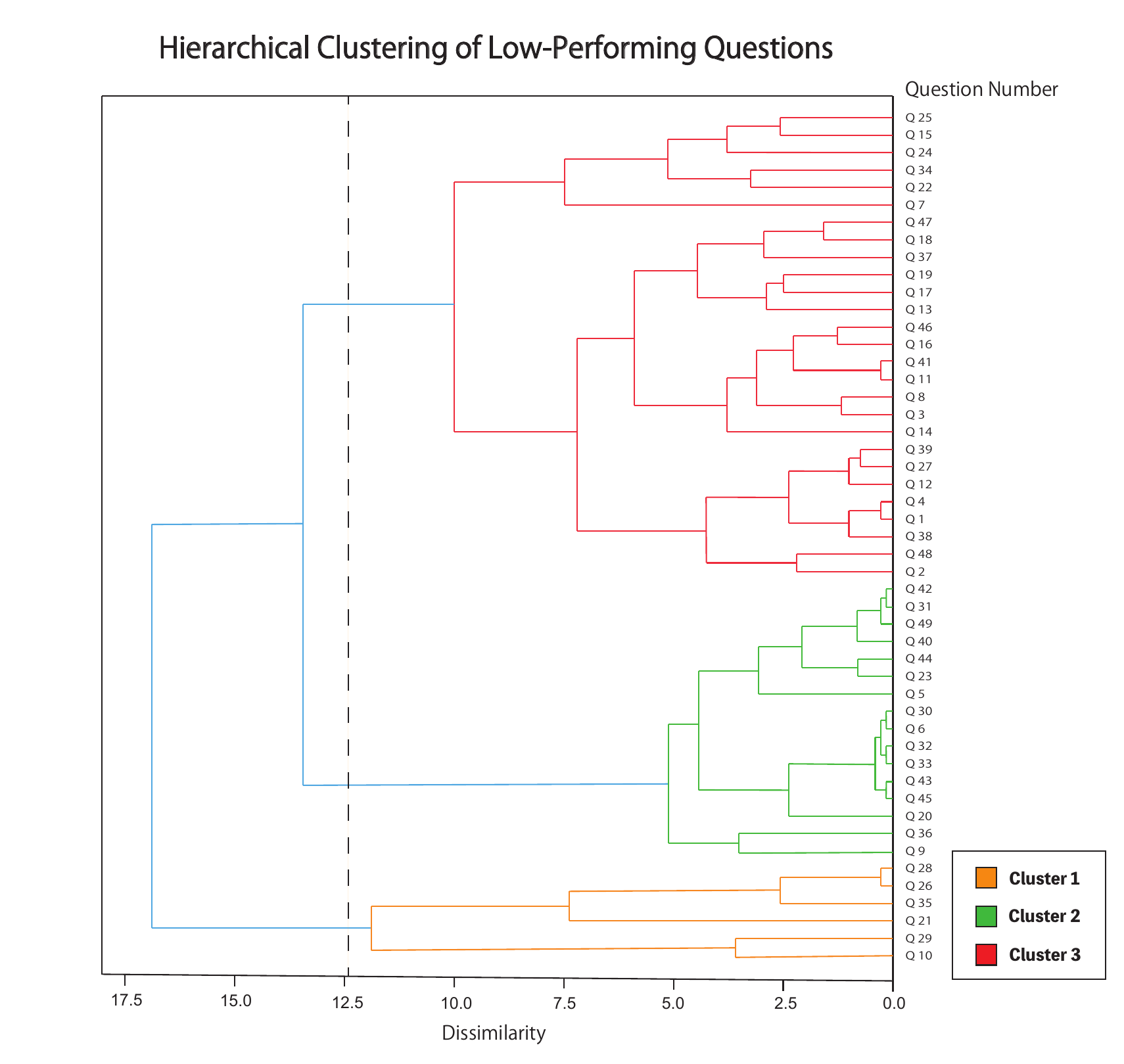}
\caption{Hierarchical Clustering of Low-Performing Questions}
\label{f13}
\end{figure}

\revc{The} error analysis examines failure patterns across graph-based RAG (Neo4j), vector-based RAG (OpenAI Playground), and BifrostRAG, focusing on multi-hop questions where one or more systems achieved precision or recall below 50\%. Of the 93 questions analyzed, 49 met this performance threshold. Hierarchical clustering is used here to group low-performing questions into three clusters that reveal distinct failure patterns across the three QA systems (Figure~\ref{f13}). Each label on the right corresponds to one of the 49 questions for which at least one system (BifrostRAG, vector-based RAG, or graph-based RAG) achieved precision or recall below 0.50. The horizontal axis (`Dissimilarity') reports the linkage distance at which questions are merged, indicating how different two questions or groups of questions are in terms of their model performance; larger values indicate that the merged clusters are more dissimilar in their performance. The vertical dashed line at a dissimilarity of 12.3 indicates the chosen cut level on the dendrogram. This threshold was selected by inspecting the dendrogram for a marked increase in linkage distance, which yields three well-separated and interpretable clusters. Orange, green, and red branches correspond to Cluster 1, Cluster 2, and Cluster 3, respectively, which represent different types of error patterns for vector-based RAG, graph-based RAG, and BifrostRAG as discussed in the following analysis.

Cluster 1 contains 6 questions where both graph-based RAG and BifrostRAG perform poorly. These questions typically feature short queries with frequently occurring terminology. For example, ``True or False: Fall protection systems may be secured to the scaffold itself'' contains common entities that appear across numerous regulatory subparts: ``fall protection system'' and ``scaffold.'' Both graph-based Cypher queries and BifrostRAG's entity and triple matching generate overly broad candidate pools, making relevant section identification difficult. Vector-based RAG performs significantly better by leveraging semantic similarity to capture question intent when common terminology creates retrieval noise.

Cluster 2 consists of 16 questions where only graph-based RAG performs poorly. These questions involve terminologies requiring contextual understanding for correct interpretation. For instance, when asked about ``safety precautions for working on a steep roof with areas of rot,'' the graph-based RAG fails to connect this phrasing with structural integrity provisions, instead incorrectly returning sections about ``wear'' of safety equipment. This demonstrates the system's reliance on entity matching rather than contextual comprehension. To confirm this observation, \revc{the LangChain framework integrated into Neo4j's RAG system  is tested separately by deploying its GraphCypherQAChain locally}. \revc{The} findings indicated that although the generated Cypher queries were generally accurate, the system struggled to locate relevant sections due to the requirement of exact syntactic matching for target entities and relationships. In contrast, BifrostRAG demonstrates superior performance through semantic matching of triple representations, enabling accurate interpretation of ambiguous terminology within regulatory contexts.

Cluster 3 comprises 27 questions where BifrostRAG significantly outperforms both alternatives, particularly in recall. These questions require comprehensive information gathering across document sections. Vector-based RAG identifies relevant information but misses additional content due to limited retrieval breadth. Three key limitations contribute to this gap: inability to follow explicit cross-references through semantic similarity alone, context loss when text chunking separates related sections, and diluted term representations in lengthy sentences where specific meanings become averaged across broader contexts. \revc{The proposed} hybrid retrieval mechanism addresses these limitations by combining graph-based hierarchical document representation with semantic understanding through entity and triple representations, enabling the system to follow regulatory connections while maintaining semantic comprehension.

\subsection{Ablation Study and Threshold Sensitivity Analysis}
To thoroughly evaluate the contributions of individual components to overall system performance, \revc{an ablation study and a threshold sensitivity analysis were conducted.} BifrostRAG employs a dual-graph framework comprising ENG and DNG. The ablation study isolates the impact of each graph by examining performance under different configurations. \revc{The ENG-only setup is adopted as the baseline,} as it forms the system’s core retrieval structure. The DNG is designed to enhance ENG retrieval by capturing document-level navigational relationships. The sensitivity analysis investigates the effect of varying similarity thresholds in the iEntity Refiner (IR), the component responsible for merging semantically similar entities. \revc{Five threshold levels are tested:} $\tau$= 80\%, 85\%, 90\%, 95\%, and 100\%. A threshold of 100\% restricts merging to exact matches, while lower thresholds allow merging based on semantic similarity. Both experiments are conducted using 44 multi-hop questions drawn from subparts P and M.

As shown in Table \ref{t13}, augmenting ENG with DNG leads to significant performance gains across all three evaluation metrics. Most notably, recall improves markedly from 63.85\% in the ENG-only setup to 84.85\% when DNG is included. These results underscore DNG’s value in supporting multi-hop reasoning by facilitating more thorough retrieval of relevant content, particularly by surfacing explicit relationships between provisions in safety regulations. The threshold sensitivity analysis reveals clear performance patterns. With all other parameters held constant, the system achieves peak F1 performance at $\tau$= 95\%. This indicates that the semantic merging provided by the IR component is both necessary and effective, but must be applied judiciously. A strict 100\% threshold results in fragmented or redundant entity representations, impairing the model's ability to integrate related information. In contrast, thresholds below 90\% lead to overly permissive merging, conflating distinct entities and thereby reducing both precision and recall by introducing irrelevant or misleading content.

\begin{table}[!ht]
\centering
\footnotesize
\renewcommand{\arraystretch}{1.1}
\linespread{1.1}
\selectfont
\caption{Ablation and Sensitivity Analysis Results}
\label{t13}
\begin{tabular}{lccc}
\toprule
 \textbf{Model Configuration} & \textbf{Precision} & \textbf{Recall} & \textbf{F1} \\
\midrule
 Baseline 1: ENG only, IR ( $\tau=100 \%$ ) & 86.18\% & 63.45\% & 68.38\% \\
 Baseline 2: ENG only, $\operatorname{IR}(\tau=95 \%)$ & 87.65\% & 66.89\% & 72.39\% \\
 Baseline 3: ENG only, IR ( $\tau=90 \%$ ) & 87.47\% & 65.93\% & 71.95\% \\
 Baseline 4: ENG only, IR ( $\tau=85 \%$ ) & 83.38\% & 62.48\% & 65.71\% \\
 Baseline 5: ENG only, IR ( $\tau=80 \%$ ) & 81.76\% & 60.52\% & 64.41\% \\
 Full 1: ENG + DNG, IR ( $\tau=100 \%$ ) & 91.18\% & 84.42\% & 86.58\% \\
 Full 2: $\mathrm{ENG}+\mathrm{DNG}, \mathrm{IR}(\tau=95 \%)$ & 93.17\% & 86.26\% & 89.90\% \\
 Full 3: $\mathrm{ENG}+\mathrm{DNG}, \mathrm{IR}(\tau=90 \%)$ & 92.76\% & 87.53\% & 89.05\% \\
 Full 4: ENG + DNG, IR ( $\tau=85 \%$ ) & 90.51\% & 84.48\% & 85.18\% \\
 Full 5: ENG + DNG, IR ( $\tau=80 \%$ ) & 88.57\% & 81.55\% & 83.89\% \\
\bottomrule
\end{tabular}
\end{table}

\section{Discussion}
This \revc{paper} demonstrates the efficacy of a RAG framework for answering complex, multi-hop questions from safety regulations. \revc{The} central contribution is an end-to-end automated pipeline that translates raw regulations into two complementary knowledge graphs, enabling both graph traversal and semantic retrieval. This hybrid mechanism proves effective in providing LLMs with comprehensive, contextually relevant information for generating grounded answers.

\subsection{Comparison with State-of-the-Art Solutions}
When evaluated against a test set of 93 multi-hop questions, \revc{the} proposed system significantly outperforms leading end-to-end solutions, including OpenAI's vector-based RAG and Neo4j's graph-based RAG approaches. The most pronounced advantages emerge in recall performance, demonstrating \revc{the} system's superior capability to retrieve the complete regulatory information required for accurate QA. Error analysis reveals fundamental limitations inherent in general-purpose systems when applied to domain-specific regulatory contexts. Neo4j's approach struggles because it is not designed to interpret or reason over complex document structures. Another limitation is its reliance on entity identification to initiate graph traversal. However, entities referenced in questions may not be identical with those in the regulations \citep{pujara2017sparsity} and regulations also frequently employ multiple expressions to refer to identical entities \citep{wang2023deep}. This ambiguity necessitates interpreting broader semantic context to correctly identify key entities--a capability that Neo4j's Cypher-based query approach alone cannot achieve \citep{omar2023chatgpt,tan2023chatgpt}. At a more fundamental level, Neo4j's underperformance in \revc{the} experiments reflects a mismatch between the strengths of conventional graph-based RAG systems and the unique demands of regulatory reasoning. Traditional graph-based RAG excels in domains with clean, semantically structured knowledge graphs, such as inferring relational chains like ``A is B's mother, B is C's father; therefore, A is C's grandmother.'' However, as \revc{the} results demonstrate, it struggles with the more complex multi-hop questions characteristic of regulatory contexts. These questions require navigating both the formal structure of regulatory documents and the implicit relationships embedded in ambiguous legal language. Such challenges necessitate richer semantic understanding and more flexible retrieval mechanisms than conventional graph-based approaches provide.

In contrast, \revc{the proposed} system's performance advantage derives from two key innovations: a dual-knowledge graph architecture and a hybrid retrieval mechanism. The architecture employs two distinct yet complementary graphs: the Document Navigator Graph maps formal regulatory structure, including hierarchical relationships and cross-references, while the Entity Network Graph links sections sharing semantically similar entities or triples. \revc{The proposed} hybrid retrieval mechanism leverages this dual architecture through structured Cypher queries that navigate the Document Navigator Graph to follow formal structure, combined with vector-based similarity searches that query the Entity Network Graph to uncover semantically-linked information across the corpus. This integrated approach effectively bridges fragmented regulatory texts, enabling comprehensive information retrieval and accurate multi-hop QA.

\subsection{Practical implications}
The methodology developed in this \revc{paper} directly addresses a critical responsible AI challenge in the modern construction industry. As the sector increasingly deploys LLMs for automated compliance verification \citep{adil2025using} and virtual instruction agent \citep{sabir2025personalized}, incomplete information retrieval creates acute operational risks, particularly given that general-purpose RAG systems prove ill-suited to safety documents' linguistic and structural complexities. While earlier studies have acknowledged these challenges \citep{wu2025retrieval,lee2024performance,chen2024knowledge}, \revc{this paper} provides one of the first systematic approaches to mapping and resolving complex inter-dependencies between regulatory provisions. By bridging the gap between current AI capabilities and stringent safety requirements, this \revc{paper} establishes a foundation for the digital transformation of construction safety management.

\revc{This paper} also contributes important insights for incremental knowledge graph construction, specifically regarding coreference resolution--the process of identifying and merging heterogeneous linguistic expressions that refer to the same entity across various data sources. Without effective coreference resolution, knowledge graphs become fragmented and inconsistent, leading to misleading retrieval outcomes \citep{lairgi2025text2kg}. This approach provides actionable guidance for practitioners constructing high-quality knowledge graphs for construction safety regulations.

\subsection{Contributions to Methodological Discussions}
Beyond immediate technological contributions, this \revc{paper} advances methodological discussions about RAG creation in the era of generative AI with broad impact extending beyond safety applications. Methodologically, most existing RAG research focuses on model fine-tuning, algorithmic refinement, and other micro-level optimizations \citep{xue2024question,lee2024performance,jeon2025hybrid,zhang2025knowledge,zheng2025automating,chen2024knowledge}. What distinguishes \revc{the proposed} approach is its macro-level, ``architecture-first'' methodology for RAG framework development. This methodology began with systematic analysis of document complexity, with these insights directly informing architectural design. In particular, \revc{a} detailed examination of the characteristics of safety regulation texts illuminates why conventional RAG approaches underperform in this context. By contrast, \revc{a} specialized architecture was explicitly designed to overcome these limitations through dual-graph reasoning and hybrid retrieval mechanism. The final system validated in this \revc{paper} reaffirms that human domain expertise remains essential when building and scaffolding specialized AI systems. \revc{The proposed method} could inspire future research in developing domain-specific RAG applications to address sector-specific challenges.

In addition, \revc{LLMs were integrated throughout the} automated pipeline using ontology-free entity extraction and zero-shot prompt engineering without requiring sophisticated fine-tuning. \revc{The} prompting strategy employs iterative refinement \citep{jeon2025hybrid}: starting with simple prompts, evaluating outputs, and progressively adding specific guidelines to shape extraction behavior. \revc{The prompts were refined on one regulatory subpart (achieving an F1-score of 93.6\% in Subpart P) and then applied to other subparts, where consistent performance was maintained (84.1\% in L, 81.6\% in M, 95\% in X)}. This demonstrates promising generalizability despite varying organizational structures, logic flows, and distinct terminologies across different subparts \citep{wang2023deep}. \revc{While the value of formal ontologies and model fine-tuning} is acknowledged in many applications, \revc{the} findings suggest that this ontology-free, zero-shot approach represents a promising alternative. In particular, it offers an accessible pathway for even non-technical domain experts to build graph-based RAG systems.

Finally, \revc{this paper} challenges the standard reliance on formal graph languages like Cypher for retrieval graph-based RAG systems by introducing a hybrid retrieval mechanism. Cypher requires exact syntactic and semantic matches between queries and schemas \citep{francis2018cypher}. While Neo4j integrates translation tools like LangChain's GraphCypherQAChain that may lower technical barriers in crafting Cypher queries, \revc{the} results demonstrate that these translations still struggle with complex safety regulations. This observation aligns with findings regarding other similar translation tools \citep{meyer2024assessing}. \revc{The proposed} hybrid approach combines two retrieval modalities: \revc{Cypher is employed} exclusively for the well-structured Document Navigator Graph while \revc{embedding similarity is used for} the unstructured Entity Network Graph. This approach proves more fault-tolerant with natural language questions and more adequately handles entity ambiguities in safety regulation texts. Through comparative evaluation and error analysis of three approaches, \revc{this paper} provides methodological insights into optimizing retrieval-structure alignment for effective information identification.

\subsection{Limitations and Future Research}
Several limitations of this \revc{paper} provide directions for future research. First, this \revc{paper} focuses exclusively on OSHA 1926, leaving transferability to other regulatory domains or organizational documents unvalidated. Second, future research should explore alternative methodologies, such as agentic approaches, to continuously enhance RAG's capability to handle multi-hop questions that are essential for LLMs' real-world applications \citep{cui2025beyond}. Third, this \revc{paper} did not directly compare \revc{the} LLM-based knowledge graph construction method with alternative approaches such as deep learning-based methods. Future work should incorporate established knowledge graph quality metrics and systematically analyze error propagation patterns while investigating methods to further improve LLMs' or other automated approaches' ability to extract entities and relationships. Finally, although \revc{the proposed} knowledge graphs already contain more nodes and edges than most existing studies \citep{zhang2025knowledge,zheng2025automating,cui2025beyond,wu2025retrieval,zhang2025dynamic}, the impact of graph scale on retrieval efficiency and QA quality requires investigation. Additionally, increasing the number of test questions would further validate the stability and consistency of \revc{the findings of this paper}. Understanding these scaling dynamics is essential for developing deployment-ready systems capable of handling enterprise-scale regulatory documents in real-world applications.

\section{Conclusions}
\revc{The linguistic and structural complexity of construction safety regulations poses significant challenges for building effective information retrieval and QA systems foundational to many automated safety management solutions. This paper introduced an end-to-end RAG methodology that converts unstructured safety regulations into knowledge graphs to enable more effective multi-hop QA using LLMs. The proposed system outperforms state-of-the-art baselines, particularly achieving better performance on complex queries that require collecting dispersed evidence. This improvement is attributed to the dual-knowledge graph architecture and the hybrid retrieval mechanism incorporated into the RAG pipeline. These results show that semantic matching and structural navigation exhibit a synergistic effect, yielding performance gains beyond those achievable by either strategy alone.}

\revc{From a technical perspective, the proposed approach addresses a critical gap in designing domain-specific RAG systems that balance advanced AI capabilities with strict safety and compliance demands. Beyond the technical contributions, this work advances methodological discussions on the synergistic integration of LLMs and knowledge graphs. Specifically, it highlights how domain knowledge informs high-level RAG architecture design, supports ontology-free and zero-shot construction of knowledge graphs, and guides the selection of retrieval strategies suited to graph structures.}

\revc{The findings carry practical implications for responsible AI deployment in construction safety contexts, where incomplete or imprecise information retrieval can lead to serious operational risks. The proposed approach can enhance safety training, support on-site real-time queries, and automate compliance verification throughout the project lifecycle. Moreover, the system’s answers are traceable to specific regulatory provisions, enabling efficient updates and ongoing maintenance as regulations evolve.}

\revc{Overall, this research advances AI-enhanced construction safety knowledge management, contributing to a safer and more efficient future for the industry. Future work should explore the generalizability of the methodology beyond OSHA 1926 to broader regulatory documents such as building codes and technical specifications. In parallel, research into more adaptive RAG paradigms is needed to further improve multi-hop QA performance in dynamic, real-world scenarios. Finally, systematic studies are essential to examine how scaling graph size and corpus complexity affect retrieval efficiency and QA quality.}

\section*{Declaration of generative AI and AI-assisted technologies in the writing process}
During the preparation of this work the authors used ChatGPT in order to improve language. After using this tool, the authors reviewed and edited the content as needed and take full responsibility for the content of the published article.

\nocite{*}
\bibliographystyle{elsarticle-num}
\bibliography{refs}
\end{document}